\documentclass[final]{cvpr}

\usepackage{times}
\usepackage{epsfig}
\usepackage{amsmath,amsfonts,bm}
\usepackage{url}
\usepackage{booktabs}
\usepackage{amssymb}
\usepackage{amsthm}
\usepackage{soul}
\usepackage{graphicx}
\usepackage{relsize}
\usepackage{float}
\usepackage{enumitem}
\usepackage{mathtools}
\usepackage{abraces}
\usepackage{xspace}

\usepackage[pagebackref=true,breaklinks=true,colorlinks=true,bookmarks=false]{hyperref}
\usepackage{cleveref}

\DeclareMathOperator*{\argmin}{arg\,min}

\newcommand{\T}{\mathcal{T}}
\newcommand{\X}{\mathcal{X}}
\newcommand{\Y}{\mathcal{Y}}

\newcommand{\E}{\mathbb{E}}
\newcommand{\R}{\mathbf{\mathbb{R}}}
\newcommand{\A}{\mathcal{A}}

\newcommand{\D}{\mathcal{D}}

\newcommand{\w}{\mathbf{w}}
\renewcommand{\vs}{\mathbf{v}}

\newcommand{\x}{\mathbf{x}}
\newcommand{\y}{\mathbf{y}}
\newcommand{\z}{\mathbf{z}}

\newcommand{\wremain}{\w^{*}_{\Dtrr}}
\newcommand{\wfull}{\w^{*}_{\D}}

\newcommand{\N}{\mathcal{N}}

\theoremstyle{plain}
\newtheorem{theorem}{Theorem}
\newtheorem{lemma}{Lemma}
\theoremstyle{definition}
\newtheorem{definition}{Definition}

\newcommand{\KL}[2]{\operatorname{KL}\big({\textstyle{#1}\,\|\,{#2}}\big)}

\newcommand{\Dtrr}{\mathcal{D}_{r}}
\newcommand{\Dtrff}{\mathcal{D}_{f}}
\newcommand{\Dte}{\mathcal{D}_{\text{Test}}}
\newcommand{\citep}{\cite}

\newcommand{\overtext}[2]{\overbrace{#1}^{\text{#2}}}
\newcommand{\undertext}[2]{\underbrace{#1}_{\text{#2}}}

\newcommand{\methodlong}{Mixed-Linear Forgetting\xspace}
\newcommand{\method}{ML-Forgetting\xspace}

\newcommand{\notea}[1]{{\color{black}#1}}
\newcommand{\note}[1]{{\color{red}#1}}
\renewcommand{\note}[1]{}
\newcommand{\cut}[1]{}

\DeclareMathOperator{\softmax}{softmax}

\def\cvprPaperID{2988} %
\def\confYear{CVPR 2021}
\begin{document}

\title{Mixed-Privacy Forgetting in Deep Networks}

\author{Aditya Golatkar$^{2,1}$ \ \ Alessandro Achille$^1$ \ \  Avinash Ravichandran$^1$ \ \ Marzia Polito$^1$ \ \ Stefano Soatto$^1$\\
$^1$Amazon Web Services \ \ \ $^2$UCLA\\
\texttt{\{aachille,ravinash,mpolito,soattos\}@amazon.com aditya29@cs.ucla.edu}
}

\maketitle

\begin{abstract}
We show that the influence of a subset of the training samples can be removed -- or ``forgotten'' -- from the weights of a network trained on large-scale image classification tasks, and we provide strong computable bounds on the amount of remaining information after forgetting. Inspired by real-world applications of forgetting techniques, we introduce a novel notion of forgetting in mixed-privacy setting, where we know that a ``core'' subset of the training samples does not need to be forgotten. While this variation of the problem is conceptually simple, we show that working in this setting significantly improves the accuracy and guarantees of forgetting methods applied to vision classification tasks. Moreover, our method allows efficient removal of all information contained in non-core data by simply setting to zero a subset of the weights with minimal loss in performance. We achieve these results by replacing a standard deep network with a suitable linear approximation. With opportune changes to the network architecture and training procedure, we show that such linear approximation achieves comparable performance to the original network and that the forgetting problem becomes quadratic and can be solved efficiently even for large models. Unlike previous forgetting methods on deep networks, ours can achieve close to the state-of-the-art accuracy on large scale vision tasks. In particular, we show that our method allows forgetting without having to trade off the model accuracy.
\end{abstract}

\note{
\begin{enumerate}
    \item Make it clear that the quadratic problem has an exact solution but we need to solve it approximately since it is too large
    \item Make it clear that the solution we find is approximate, the quality of the solution depnds on the number of steps. Figure 4 shows the dependency of remaining information on steps and Theorem 1 shows it theoretically
\end{enumerate}
}

\section{Introduction}

When building a classification system, one rarely has all the data to be used for training available at the outset. More often, one starts by pre-training a model with some ``core'' dataset (e.g. ImageNet, or datasets close to the target task) and then incorporates various cohorts of task-specific data as they become available from diverse sources. In some cases, the wrong data may be incorporated inadvertently, or the owners may change their mind and demand that their data be removed. One can, of course, restart the training from scratch every time such a demand is made, but at a significant cost of time and disruption. What if one could remove the effect of cohort(s) of data {\em a-la-carte}, without re-training, in a way that the resulting model is  functionally indistinguishable from one that has never seen the cohort(s) in question, and in addition has  no residual information about it buried in the weights of the model?  Of course, forgetting can always be trivially achieved by zeroing the weights or replacing them with random noise, but this comes at the expense of the accuracy of the model. Can we forget the cohort of interest without interfering with information about the other data and preserving, to the extent possible, the accuracy of the trained model? Recently, the problem of forgetting has received considerable attention \cite{golatkar2019eternal,golatkar2020forgetting,ginart2019making,guo2019certified,baumhauer2020machine,izzo2020approximate,neel2020descenttodelete, wu2020deltagrad,bourtoule2019machine, sommer2020towards, garg2020formalizing,brophy2020dart,nguyen2020variational}, but solutions have focused on simpler machine learning problems such as linear logistic regression. Removing information from the weights of a standard convolutional network still remains an open problem, with some initial results working only on small scale problems \cite{golatkar2019eternal,golatkar2020forgetting}. This is mainly due to the highly non-convex loss-landscape of CNNs, which makes the influence of a particular sample on the optimization trajectory and the final weights highly non-trivial to model.

In this paper we introduce \methodlong (\method), a method to train large scale computer vision models in such a way that information about a subset of the data can be removed on request -- with strong bounds on the amount of remaining information -- while at the same time retaining close to the state of the art accuracy on the tasks. To the best of our knowledge, this is the first algorithm to achieve forgetting for deep networks trained on large-scale computer vision problems without compromising the accuracy. To further improve the performance in realistic use-cases, \note{I would reframe this: Here you make it sound like you are concocting a scenario just to improve performance, and cite the  main novelty as allowing forgetting without loss of precision. I would not mention the second (presumably every paper on the topic makes that claim) and focus on the first by turning it backwards: You tackle the problem of forgetting in a mixed-privacy setting that is better aligned with real use cases, where a ``core'' dataset rarely if ever needs to be forgotten (and if it does, given the scale, a complete re-train is justified) whereas small cohorts incorporated in fine-tuning may need to be forgotten on-demand. For this problem, you show that you can forget the cohort, give bounds on the residual information, an retain near-original performance, understood as performance that would have been obtained by training with all the data but the one to be forgotten.} we introduce the notion of forgetting in a \textit{mixed-privacy setting},  that is, when we know that a subset $\D_c \subset \D$ of the training dataset, which we call \emph{core data},  will not need to be forgotten. For example, the core data may be a large dataset of generic data used for pre-training (e.g., ImageNet) or a large freely available collection of task-specific data (e.g., a self-driving dataset) which is not likely subject to changes.
We show that \method can naturally take advantage of this setting, to improve both accuracy and bounds on the amount of remaining information after forgetting. %

One of the main challenges of forgetting in deep networks is how to estimate the effects of a given training sample on the parameters of the model, which has lead the research to focus on simpler convex learning problem such as linear or logistic regression, for which a theoretical analysis is feasible. %
To address this problem, \methodlong uses a first-order Taylor-series inspired decomposition of the network to learn two sets of weights: a core set $\w_c$ which is trained only with the core data $\D_c$ using a standard (non-convex) algorithm, and a set of linear $\w$ of user weights, which is trained to minimize a quadratic loss function on the changeable user data $\D$.
The core weights are learned through standard training (since forgetting is not required on core data), while the user weights are obtained as the solution to a strongly convex quadratic optimization problem.
This allows us to remove influence of a subset of the data with strong guarantees. Moreover, by construction, simply setting to zero the user weights removes influence of all changeable data with the lowest possible drop in performance, thus easily allowing the user to remove all of their data at the same time.

\vspace{.5em}

\noindent To summarize, our key contributions are:
\begin{enumerate}[labelwidth=!, labelindent=0pt]
    \item We introduce the problem of forgetting (unlearning or data deletion or scrubbing) in a mixed-privacy setting which, compared to previous formalizations, is better taylored to standard practice, and allows for better privacy guarantees.
    \item In this setting we propose \method. \method trains a set of non-linear \textit{core} weights and a set of linear \textit{user} weights, which allow it to achieve both good accuracy, thanks to the flexibility of the non-linear weights, and strong privacy guarantees thanks to the linear weights.%
    \item As a side effect, all the user data may be forgotten completely with the lowest possible drop in performance by simply erasing the user weights.
    \item We show that \method can be applied to large-scale vision datasets, and enjoys both strong forgetting guarantees and test time accuracy comparable to standard training of a Deep Neural Network (DNN). To the best of our knowledge, this is the first forgetting algorithm to do so. 
    \item Furthermore, we show that \method can handle multiple sequential forgetting requests without degrading its performance, which is important for real world applications. %
\end{enumerate}

\section{Related Work}

\textbf{Forgetting.} The problem of machine unlearning is introduced in \cite{cao2015towards} as an efficient forgetting algorithm for statistical query learning. \cite{mirzasoleiman2017deletion,ginart2019making} gives method for forgetting for particular class of learning algorithms, such as k-means clustering. Other methods involve splitting the data into multiple subsets and train models separately on combinations of them \cite{bourtoule2019machine,wu2020deltagrad}. This allows perfect forgetting, but incurs in heavy storage costs as multiple models/gradients need to be stored. In the context of model interpretability and cross-validation, \cite{koh2017understanding,giordano2019swiss} provided a hessian based method for estimating the influence of a training point on the model predictions.
\cite{baumhauer2020machine} proposed a method for hide information about an entire class from the output logits, but does not remove information from the model weights.
\cite{guo2019certified} proposed to remove information from the weights on convex problems using Newton's method, and uses differential privacy \cite{abadi2016deep,dwork2014algorithmic,chaudhuri2011differentially,chaudhuri2008privacy} to certify data removal.
\cite{izzo2020approximate} provides a projective residual update method using synthetic data points to delete data points from linear/logistic regression based models.  \cite{nguyen2020variational} proposed an unlearning mechanism for logistic regression and gaussian processes in a Bayesian setting using variational inference.
Recently, \cite{neel2020descenttodelete} proposed a gradient descent based method for data deletion in convex settings, with theoretical guarantees for multiple forgetting requests. They also introduce the notion of statistical indistinguishably of the entire state or just the outputs similar to the information theoretic framework of \cite{golatkar2020forgetting}. We use some of their proof techniques for our theoretical results.

Deep Networks provide additional challenges to forgetting due to their highly non-convex loss functions. \cite{golatkar2019eternal} proposed an information theoretic procedure to scrub the information from intermediate layers of DNN trained with stochastic gradient descent (SGD), exploiting the stability of SGD \cite{hardt2015train}. %
They also bound the amount of remaining information in the weights \cite{achille2019where} after scrubbing. \cite{golatkar2020forgetting} extend the framework of \cite{golatkar2019eternal} to activations. They also show that an approximation of the training process based on a first-order Taylor expansion of the network (NTK theory) can be used to the estimate the weights after forgetting.
This approximation works well on small scale vision datasets. However, the approximation accuracy and the computational cost degrade for larger datasets (in particular the cost is quadratic in the number of samples). We also use linearizarion but, crucially, instead of linearly approximating the training dynamics of a non-linear network, we show that we can directly train a linearized network for forgetting. This ensures that our forgetting procedure is correct, and it allows us to scale easily to standard real-world vision datasets.

\textbf{Linearization.} Using a first-order Taylor expansion (linearization) of the network to study its behavior has gained interest recently in the NTK theory \cite{jacot2018neural,li2019enhanced} as a tool to study the dynamics of DNNs in the limit of infinite filters. \cite{Mu2020Gradients} shows that aside from a theoretical tool, it is possible to directly train a (finite) linearized network using an efficient algorithm for the Jacobian-Vector product computation. \cite{achille2020lqf} show that with some changes to the architecture and training process, linearized models can match the performance of non-linear models on many vision tasks, while still maintaining a convex loss function.

\section{Preliminaries and Notations}
We use the empirical risk minimization (ERM) framework throughout this paper for training. Let $\D = \{\x_i, \y_i\}_{i=1}^{n}$ be a dataset where $\x_i \in \X$ denotes the input datum (for example, images) and $\y_i \in \Y$ the corresponding output (for example, one hot vector in classification). Given an input image $\x$, let $f_{\w}(\x): \X \times \R^d \rightarrow \Y$ (for instance, a DNN) be a function parameterized by $\w \in \R^d$ used to model the relation $\X \rightarrow \Y$. Given a input-target pair $(\x, \y) \in (\X, \Y)$, we denote empirical risk or the training loss for $(\x, \y)$ by $\ell(f_\w(\x),\y): \X \times \Y \times \R^d \rightarrow \R^+$. We will sometimes abuse notation and use $\ell(\w)$ by dropping $\x, \y$. For a training dataset $\D = \{\x_i,\y_i\}_{i=1}^{n} \subset (\X, \Y)^n$, we denote the empirical risk/total training loss on $\D$ by $L_{\D}(\w) \triangleq \frac{1}{|\D|} \sum_{i \in \D} \ell(f_\w(\x_i), \y_i)$. We will interchangebly use $L_{\D}(f_{\w})$ with $L_{\D}(\w)$. Let $\A_{\tau}(L_{\D}(\w_0)): (\X, \Y)^{n} \times \R^d \times \mathbb{N} \rightarrow \R^d$, denote the weights obtained after $\tau$ steps of a training algorithm $\A$ using $\w_0$ as the initialization (for examples, SGD  in our case). 
We denote with $\|\w\|$ the $L_2$ norm of a vector $\w$ and with $\lambda_{\text{Max}}(Q)$  the largest eigenvalue of a matrix Q. To keep the notation uncluttered, we also use the shorthand $\nabla_{\w} L(\w') \triangleq \nabla_{\w} L(\w)|_{\w=\w'}$.

\section{The Forgetting Problem}
\label{sec:forgetting}

The weights $\w$ of a trained deep network $f_{\w}(\x)$ are a (possibly stochastic) function of the training data $\D$. As such, they may retain information about the training samples which an attacker\footnote{An attacker is any agent intent to extract information about the data used for training.} can extract from knowledge of the weights or outputs at inference time. A forgetting procedure is a function $S(\w, \D, \D_F)$\footnote{We will abuse the notation and write $S(\w)$ when its arguments are clear from the context.} (also called scrubbing function) which, given a set of weights $\w$ trained on $\D$ and a subset $\Dtrff \subset \D$ of images to forget, outputs a new set of weights $\w'$ which are indistinguishable from weights obtained by training without $\Dtrff$.

\textbf{Readout functions.} The success of the forgetting procedure, can be measured by looking at whether a discriminator function $R(\w)$ that can guess -- at better than chance probability --  whether a set of weights $\w$ was trained with or without $\Dtrff$ or whether it was trained with $\Dtrff$ and then scrubbed. Following \cite{golatkar2019eternal,golatkar2020forgetting} we call such functions \textit{readout functions}. A popular example of readout function is the confidence of the network (that is, the entropy of the output softmax vector) on the samples in $\Dtrff$: Since networks tend to be overconfident on their training data \cite{guo2017calibration,kristiadi2020being}, a higher than expected confidence may indicate that the network was indeed trained on $\Dtrff$. We discuss more read-out functions in \Cref{sec:readout-functions}. Alternatively, we can measure the success of the forgetting procedure by measuring the amount of remaining mutual information $\mathcal{I}(S(\w); \Dtrff)$\footnote{$\mathcal{I}(\x,\y) = \E_{P_{\x,\y}}[ \log{(P_{\x,\y}/P_{\x}P_{\y}})]$ is the mutual information between $\x$ and $\y$, where $P_{\x,\y}$ is the joint distribution and $P_{\x},P_{\y}$ are the marginal distributions.} between the scrubbed weights $S(\w)$ and the data $\Dtrff$ to be forgotten. While this is more difficult to estimate, it can be shown that $\mathcal{I}(S(\w); \Dtrff)$ upper-bounds the amount of information that any read-out function can extract \cite{golatkar2019eternal,golatkar2020forgetting}. That is, it is an upper-bound on the amount of information that an attacker can extract about $\Dtrff$ using the scrubbed weights $S(\w)$.

\textbf{Quadratic forgetting.} An important example is forgetting in a linear regression problem, which has a quadratic loss function $L_{\D}(\w) = \sum_{(\x_i, \y_i) \in \D} \|\y_i - \w^T \x_i\|^2$. Given the weights $\w = \A_{\tau}(L_{\D}(\w))$ obtained after training on $L_{\D}(\w)$ using algorithm $\A$, the optimal forgetting function is given by:
\begin{align}
\label{eq:quadratic-scrubbing}
\w \mapsto \w - H^{-1}_{\Dtrr} \nabla_{\w} L_{\Dtrr}(\w),    
\end{align}
where $H^{-1}_{\Dtrr}, \nabla_{\w} L_{\Dtrff}(\w)$ is the hessian and gradient of the loss function computed on the remaining data respectively. When $\A_{\tau}(L_{\D}(\w))=\w^* \triangleq \text{argmin}_{\w}L_{\D}(\w)$, we can replace $L_{\Dtrr}(\w^*)$ with $-L_{\Dtrff}(\w^*)$ in which case it can be interpreted as a reverse Newton-step that unlearns the data $\Dtrff$ \cite{guo2019certified,golatkar2019eternal}. Since, as we will see later, the ``user weights'' of \method minimize a similar quadratic loss function,  \cref{eq:quadratic-scrubbing} also describes the optimal forgetting procedure for our model. The main challenge for us will be how to accurately compute the forgetting step since the Hessian matrix can't be computed or stored in memory due to the high-number of parameters of a deep network (\Cref{sec:forgetting-procedure}).

\textbf{Convex forgetting.} Unfortunately, for more general machine learning models we do not have a close form expression for the optimal forgetting step. However, it can be shown \cite{koh2017understanding} that \cref{eq:quadratic-scrubbing} is always a first-order approximation of the optimal forgetting. \cite{guo2019certified} shows that for strongly convex Lipschitz loss functions, the discrepancy between \cref{eq:quadratic-scrubbing} and the optimal forgetting is bounded. Since this discrepancy -- even if bounded -- can leak information, a possible solution is to add a small amount of noise after forgetting:
\begin{align}
\label{eq:netwon-forgetting}
    \w \mapsto \w + H^{-1}_{\Dtrr}(\w) \nabla_{\w} L_{\Dtrff}(\w) + \sigma^2 \epsilon,
\end{align}
where $\epsilon \sim N(0, I)$ is a vector of random Gaussian noise, which aims to destroy any information that may leak due to small discrepancies. Increasing the variance $\sigma$ of the noise destroys more information, thus making forgetting more secure, but also reduces the accuracy of the model since the weights are increasingly random. The curve of possible Pareto-optimal trade-offs between accuracy and forgetting can be formalized with the Forgetting Lagrangian \cite{golatkar2019eternal}.

Alternatively, to forget data in a strongly convex problem, one can fine-tune the weights on the remaining data using perturbed projected-GD \cite{neel2020descenttodelete}. Since projected-GD converges to the unique minimum of a strongly convex function regardless of the initial condition (contrary to SGD, which may not converge unless proper learning rate scheduling is used), this is guaranteed to remove all influence of the initial data \cite{neel2020descenttodelete}. The downside is that gradient descent (GD) is impractical for large-scale deep learning applications compared to SGD, projection based algorithms are not popular in practice, and the commonly used loss functions are not generally Lipschitz.

\textbf{Non-convex forgetting.} 
Due to their highly non-convex loss-landscape, small changes of the training data can cause large changes in the final weights of a deep network. This makes application of \cref{eq:netwon-forgetting} challenging. \cite{golatkar2019eternal} shows that pre-training helps increasing the stability of SGD and derives a similar expression to \cref{eq:netwon-forgetting} for DNNs, and also provides a way to upper-bound the amount of remaining information in a DNN. \cite{golatkar2019eternal} builds on recent results in linear approximation of DNNs, and approximate the training path of a DNN with that of its linear approximation. While this improves the forgetting results, the approximation is still not good enough to remove all the information. Moreover, computing the forgetting step scales quadratically with the number of training samples and classes, which restricts the applicability of the algorithm to smaller datasets.

\section{\methodlong}

Let $f_{\w}(\x)$ be the output of a deep network model with weights $\w$ computed on an input image $\x$. For ease of notation, assume that the core dataset and the user dataset share the same output space (for example, the same set of classes, for a classification problem). After training a set of weights $\w_c$ on a core-dataset $\D_c$ we would like to further perturb those weights to fine-tune the network on user data $\D$. We can think of this as solving the two minimization problems:
\begin{align}
    \w_c^* &= \argmin_{\w_c} L_{\D_c}(f_{\w_c}) \label{eq:public-training}\\
    \w_u^* &= \argmin_{\w_u} L_{\D}\big({f_{\w_c^* + \w_u}}\big) \label{eq:private-training}
\end{align}
where we can think of the user weights $\w_u$ a perturbation to the core weights that adapts them to the user task. However, since the deep network $f_\w$ is not a linear function in of the weights $\w$, the loss function $L_{\D}\big({f_{\w_c^* + \w_u}}\big)$ can be highly non-convex. As discussed in the previous section, this makes forgetting difficult. However, if the perturbation $\w_u$ is small, we can hope for a linear approximation of the DNN around $\w_c^*$ to have a similar performance to fine-tuning the whole network \cite{Mu2020Gradients}, while at the same time granting us easiness of forgetting.

Motivated, by this, we introduce the following model, which we call Mixed-Linear Forgetting model (ML-model):
\begin{equation}
    f^\text{ML}_{\w_c^*, \w_u}(\x) \triangleq \hspace{-.5em}\undertext{f_{\w_c^*}(x)}{trained on $\D_c$} + \overtext{\nabla_\w f_{\w_c^*}(\x) \cdot \w_u}{train on $\D$}.
    \label{equation:lineaized-model}
\end{equation}
The model $f^\text{ML}_{\w_c^*, \w_u}(\x)$ can be seen as first-order Taylor approximation of the effect of fine-tuning the original deep network $f_{\w_c^* + \w_u}(\x)$. It has two sets of weights, a set of non-linear core weights $\w_c$, which enters the model through the non-linear network $f_{\w_c}(\x)$, and a set linear user-weights $\w_u$ which enters the model linearly. Even though the model is linear in $\w_u$, it is still a highly non-linear function of $\x$ due to the non-linear activations in $\nabla_{\w} f(\w^*_c)$.

We train the model solving two separate minimization problems:
\begin{align}
    \w_c^* &= \argmin_{\w_c} L_{\D_c}^\text{CE}(f_{\w_c}), \label{eq:public-training-linear}\\
    \w_u^* &= \argmin_{\w_u} L_{\D}^\text{MSE}\big({f^\text{ML}_{\w_c^*+\w_u}}\big).
    \label{eq:private-training-linear}
\end{align}
Eq.~(\ref{eq:public-training-linear}) is akin to pretraining the weights $\w_c$ on the core dataset $\D_c$, while \cref{eq:private-training-linear} fine-tunes the linear weights on all the data $\D$. This ensures the weights $\w_c$ will only contain information about the core dataset $\D_c$, while all information about the user data $\D$ is contained in $\w_u$.
Also note that we introduce two separate loss functions for the core and user data. To train the user weights we use a mean square error (i.e., $L_2$) loss \cite{hui2020evaluation,muthukumar2020classification,golik2013cross}:
\begin{align}
\label{eq:quadratic-loss}
    \nonumber L_\D^\text{MSE}(\w_u) &= \dfrac{1}{2|\D|} \sum_{(\x, \y) \in \D} \big\| f_{\w_c^*}(\x) + \nabla_\w f_{\w_c^*}(\x) \cdot \w_u - \y \big\|^2 \\ &+ \frac{\mu}{2}\|\w_u\|^2
\end{align}
where $\y$ is a one-hot encoding of the class label. This loss has the advantage that the weights $\w_u$ are the solution to a quadratic problem, in which case the optimal forgetting step can be written in closed form (see eq.~\ref{eq:quadratic-scrubbing}). On the other hand, since we do not need to remove any information from the the weights $\w_c$, we can train them using any loss in \cref{eq:public-training}. We pick the standard cross-entropy loss, %
although this choice is not fundamental for our method. 

\subsection{Optimizing the Mixed-Linear model}
\label{sec:optimizing}

\begin{figure}
    \centering
    \includegraphics[width=1.0\linewidth]{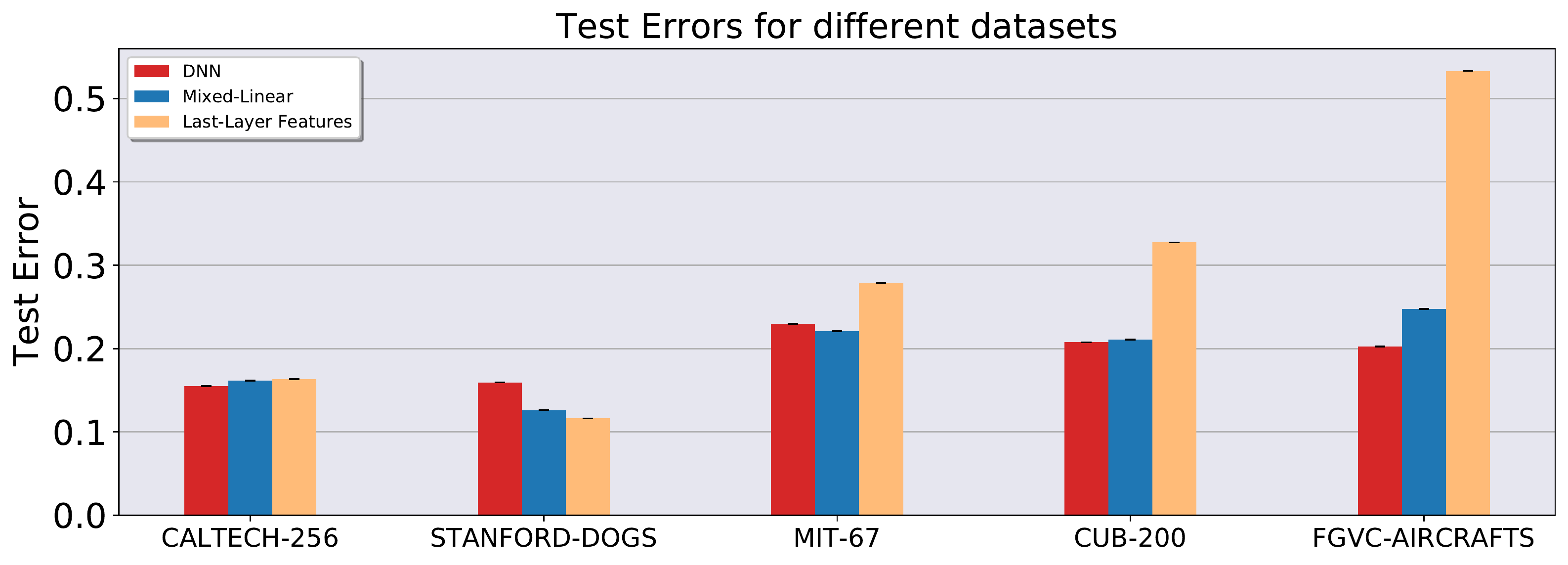}
    \caption{%
    \textbf{Mixed-Linear model has comparable accuracy to standard DNN.} Plot of the test errors for different datasets using different models. We take a ResNet-50 pretrained on ImageNet and fine-tune it using different procedures,   \textbf{(DNN)} We fine-tune the whole network on various datasets, \textbf{(Mixed-Linear)} We fine-tune the linearized ResNet-50 (\cref{equation:lineaized-model}) in a mixed private setting, \textbf{\textbf{Last-Layer Features}}: We simply fine-tune the final fully connected (FC) layer of the ResNet-50. We show that fine-tuning a linearized DNN using the mixed-privacy framework performs comparable to fine-tuning a DNN and outperforms simply fine-tuning the last FC layer.}
    \label{fig:error-comparison}
\end{figure}

Ideally, we want the ML-model to have a similar accuracy on the user data to a standard non-linear network. At the same time, we want the ML-model to perform significantly better than simply training a linear classifier on top of the last layer features of $f_{\w_c^*}$, which is the trivial baseline method to train a linear model for an object classification task. In \Cref{fig:error-comparison} (see \Cref{sec:experiments} for details) we see that this is indeed the case: while linear, the ML-model is still flexible enough to fit the data with a comparable accuracy to the fully non-linear model (DNN). However, some considerations are in order regarding how to train our ML-model. 

\textbf{Training the core model.} Eq.~(\ref{eq:public-training}) reduces to the standard training of a DNN on the dataset $\D_c$ using cross-entropy loss. We train using SGD with annealing learnig rate. In case $\D_c$ is composed of multiple datasets, for example ImageNet and a second dataset closer to the user task, we first pretrain on ImageNet, then fine-tune on the other dataset.

 \textbf{Training the Mixed-Linear model.} Training the linear weights of the Mixed-Linear model in \cref{eq:private-training} is slighly more involved, since we need to compute the Jacobian-Vector product (JVP) of $\nabla_\w f_{\w_c^*}(\x) \cdot \w_u$. While a na\"ive implementation would require a separate backward pass for each sample, \cite{Mu2020Gradients, pearlmutter1994fast} show that the JVP of a batch of samples can be computed easily for deep networks using a slightly modified forward pass. The modified forward pass has only double the computational cost of a standard forward pass, and can be further reduced by linearizing only the final layers of the network. Using the algorithm of \cite{Mu2020Gradients} to compute the model output, \cref{eq:private-training} reduces to a standard optimization, which we perform again with SGD with annealing learning rate. Note that, since the problem is quadratic, we could use more powerful quasi-Netwon methods to optimize, however we avoid that to keep the analysis simpler, since optimization speed is not the focus of this paper.

\textbf{Architecture changes.} We observe that a straightforward application of \cite{Mu2020Gradients} to a standard pre-trained ResNet-50 tend to under-perform in our setting (fine-tuning on large scale vision tasks). In particular, it achieves only slightly better performance than training a linear classifier on top of the last layer features. \notea{Following the suggetsion of \cite{achille2020lqf}, we replace the ReLUs with Leaky ReLUs, since it boosts the accuracy of linearized models.}

\section{Forgetting procedure}
\label{sec:forgetting-procedure}

The user weights $\w_u$ are obtained by minimizing the quadratic loss function $L_\D^\text{MSE}$ in \cref{eq:quadratic-loss} on the user data $\D$. Let $\Dtrff \subset \D$ denote a subset of samples we want to forget (by hypothesis $\D_c \cap \Dtrff = \emptyset$, i.e., the core data is not going to change) and let $\Dtrr = \D - \Dtrff$ denote the remaining data. As discussed in \Cref{sec:forgetting}, in case of quadratic training loss the optimal forgetting step to delete $\Dtrff$ is given by:
\begin{align}
\w_u \mapsto&\, \w_u - H^{-1}_{\Dtrr}(\w_c) g_{\Dtrr}(\w_u),
\label{eq:explicit-scrubbing}
\end{align}
where we define $g_{\Dtrr}(\w_u) \triangleq \nabla_{\w} L_{\Dtrr}(\w_u)$ and we can explicitly write the Hessian $H_{\Dtrr}(\w_c)$ of the loss \cref{eq:quadratic-loss} as:
\begin{align}
    H_{\Dtrr}(\w_c) &= \sum_{\x \in \Dtrr} \nabla_{\w} f_{\w_c}(\x)^T \nabla_{\w} f_{\w_c}(\x) + \mu I.
    \label{eq:explicit-hessian}
\end{align}
where $I$ is the identity matrix of size $d$.
Thus, forgetting amounts to computing the update step \cref{eq:explicit-scrubbing}. Unfortunately, even if we can easily write the hessian $D_{\Dtrr}$ in closed form, we cannot store it in memory and much less invert it. Instead, we now discuss how to find an approximation of the forgetting step $H^{-1}_{\Dtrr}(\w_c) g_{\Dtrr}(\w_u)$ by solving an optimization problem which does not require constructing or inverting the hessian.

Since $H_{\Dtrr}(\w_c)$ is positive definite, we can define the auxiliary loss function
\begin{align}
\hat{L}_{\Dtrr}(\vs) = \frac{1}{2} \vs^T H_{\Dtrr}(\w_c) \vs -  g_{\Dtrr}(\w_u)^T \vs \label{eq:conjugate-problem}
\end{align}
By setting the gradient to zero, it is easy to see that the forgetting update $H^{-1}_{\Dtrr}(\w_c) g_{\Dtrr}(\w_u)$ is the unique minimizer of $\hat{L}_{\Dtrr}(\vs)$, so we can recast computing the forgetting update as simply minimizing the loss $\hat{L}_{\Dtrr}(\vs)$ using SGD. In general, the product $\vs^T H_{\Dtrr}(\w_c) \vs$ of \cref{eq:conjugate-problem} can be computed efficiently without constructing the Hessian using the Hessian-Vector product algorithm \cite{koh2017understanding}. However, in our case we have a better alternative due to the fact that we use MSE loss and that ML-model is linear in weight-space: Using \cref{eq:explicit-hessian}, we have that
\begin{align}
\vs^T H_{\Dtrr}(\w_c) \vs = \sum_{\x \in \Dtrr} \|\nabla_{\w} f_{\w_c^*}(\x) \vs\|^2 + \mu \|\vs\|^2,
\end{align}
where $\nabla_{\w} f_{\w_c^*}(\x) \vs$ is a Jacobian-Vector product which can be computed efficiently (see \Cref{sec:optimizing}). Using this result, we compute the (approximate) minimizer of \cref{eq:conjugate-problem} using SGD. When optimizing \cref{eq:conjugate-problem}, we compute $g_{\Dtrr}(\w_u)$ exactly on $\Dtrr$ and approximate \cref{eq:explicit-hessian} by Monte-Carlo sampling. In \Cref{fig:info_vs_noise_compare_with_ft}, we show this method outperforms full stochastic minimization of \cref{eq:conjugate-problem}.

\textbf{Mixed-Linear Forgetting.} Let $\Delta \w_u \triangleq \A_{\tau}(\hat{L}_{\Dtrr})$ be the approximate minimizer of \cref{eq:conjugate-problem} obtained by training with SGD for $\tau$ iterations. Our Mixed-Linear (ML) forgetting procedure $S(\w)$ for the ML-model in \cref{equation:lineaized-model} is:
\begin{align}
\label{eq:ml-forgetting}
\boxed{\w_u \mapsto \w_u - \Delta \w_u + \sigma^2 \epsilon}
\end{align}
where $\epsilon \sim N(0, I)$ is a random noise vector \cite{golatkar2019eternal, golatkar2020forgetting}. As mentioned in \Cref{sec:forgetting}, we need to add noise to the weights since $\Delta \w_u$ is only an approximation of the optimal forgetting step, and the small difference may still contain information about the original data. By adding noise, we destroy the remaining information. Larger values of $\sigma$ ensure better forgetting, but can reduce the performance of the model. In the next sections, we analyze theoretically and practically the role of $\sigma$.

\textbf{Sequential forgetting.} In practical applications, we may receive several separate requests to forget the data in a sequential fashion. In such cases, we simply apply the forgetting procedure in \cref{eq:ml-forgetting} on the weights obtained at the end of the previous step. A key component is to ensure that the performance of the system does not deteriorate too much after many sequential requests, which we do  next.

\section{Bounds on Remaining Information}

\begin{figure}[t]
    \centering
    \includegraphics[width=0.85\linewidth]{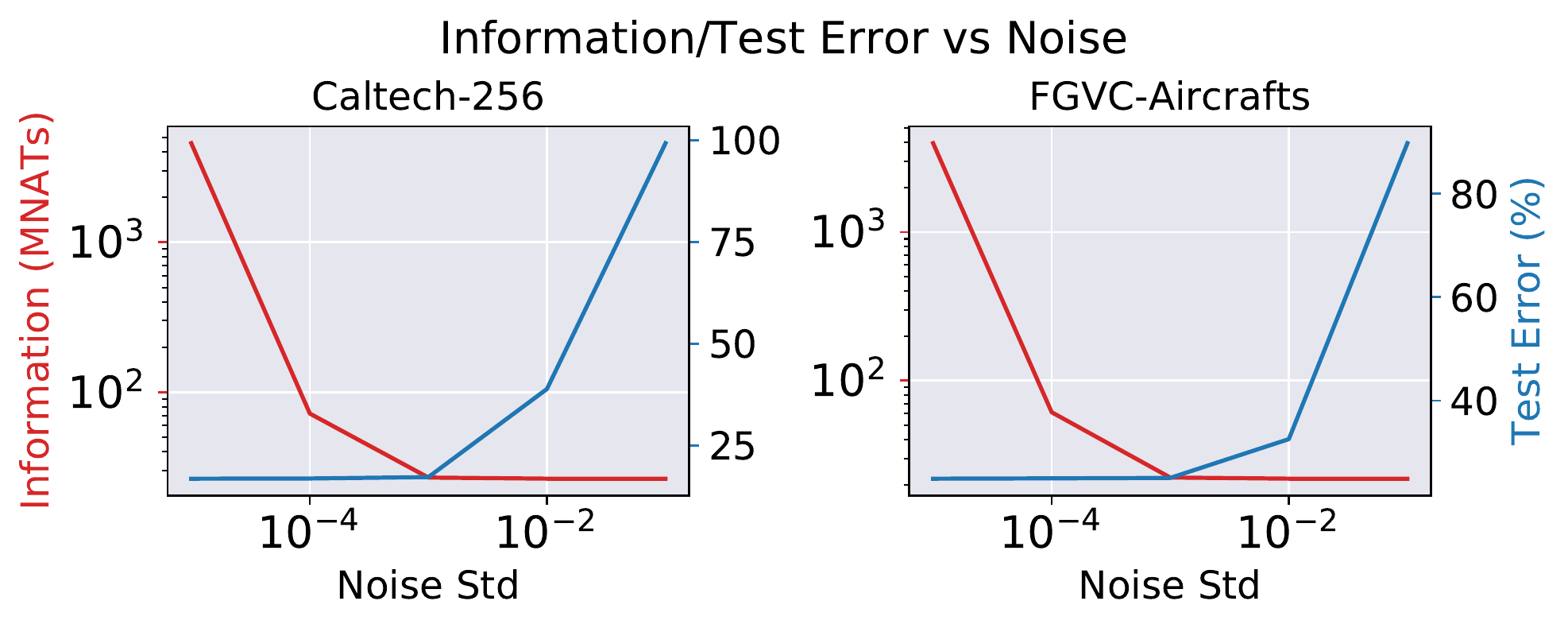}

    \includegraphics[width=0.85\linewidth]{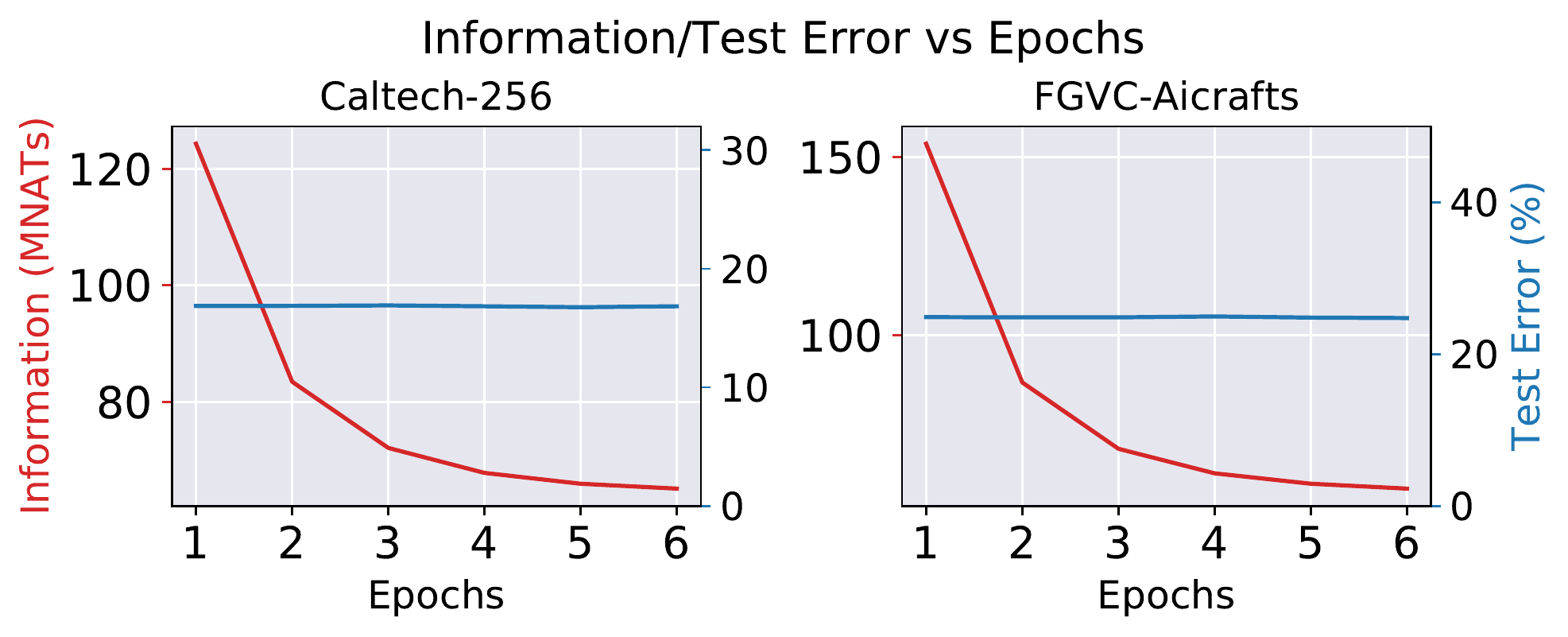}
    \caption{\textbf{Forgetting-Accuracy Trade-off} Plots of the amount of remaining information in the weights about the data to forget (red, left axis) and test error (blue, right axis) as a function of the (top) scrubbing noise and (bottom) number of optimization iterations used to compute the scrubbed weights in \cref{eq:ml-forgetting}. We aim to forget 10\% of the training data through 10 forgetting requests on the Caltech-256 (left) and Aircrafts datasets (right). Note that the remaining information in the weights decreases with an increase in the forgetting noise or the number of epochs during forgetting as predicted by the bound in \Cref{thm:information}. Increasing the forgetting noise increases the test error after forgetting (top). In terms of the computational efficiency/speed, doing 2-3 passes over the data (i.e. 2-3 epochs) is sufficient for forgetting (in terms of the test error and the remaining information) rather than re-training from scratch for 50 epochs (bottom) for each forgetting request $\Dtrff^k$. Thus providing a 16-25$\times$ speed-up per forgetting request. We fine-tune the ML-Forgetting model for 50 epochs while training the user weights. Values for $\tau$ and $\sigma$ can be chosen using these trade-off curves given a desired privacy level.}
    \label{fig:info_vs_noise_epochs}
\end{figure}

We now derive bounds on the amount of information that an attacker can extract from the weights of the model after applying the scrubbing procedure \cref{eq:ml-forgetting}. This will also guide us in selecting the optimal $\sigma$ and the number of iterations $\tau$ to approximate the forgetting step that are necessary to reach a given privacy level (see \cref{fig:info_vs_noise_epochs}).
Let $Y_{\D_F}$ denote some attribute of interest regarding $\Dtrff$ an attacker might want to access, then from Proposition 1 in \cite{golatkar2019eternal} we have:
\begin{align}
&\undertext{\mathcal{I}(Y_{\Dtrff}, S(\w))}{Recovered Information} \leq \undertext{\mathcal{I}(\Dtrff, S(\w))}{Remaining Information in Weights} \nonumber 
\end{align}
where $S(\w) = S(\w, \D, \Dtrff)$ is the scrubbing/forgetting method which given weights $\w$ trained on $\D$ removes information about $\Dtrff$ (which in our case is given by eq.~\ref{eq:ml-forgetting}). Hence, bounding the amount of information about $\Dtrff$ that remains in the weights $S(\w)$ after forgetting uniformly bounds all the information that an attacker can extract.%

We now upper-bound the remaining information $\mathcal{I}(\Dtrff, S(\w))$  after applying the forgetting procedure in \cref{eq:ml-forgetting} to our ML-model, over multiple forgetting requests. Let $\cup_{k=1}^{K} {\Dtrff}^{k}$ be the total data asked to be forgotten at the end of $K$ forgetting requests and let $\w_u^{K}$ be the weights obtained using the forgetting procedure in \cref{eq:ml-forgetting} sequentially. Then we seek to provide a bound on the mutual information between the two, i.e., $\mathcal{I} (\cup_{k=1}^{K} \Dtrff^k) \triangleq \mathcal{I}(\cup_{k=1}^{K} \Dtrff^k, \w_u^{K})$. We prove the following theorem.

\begin{theorem}[Informal]
\label{thm:information}
Let $\Delta \w_u = \A_\tau(\hat{L}_{\Dtrr})$ be the approximate update step obtained minimizing $\hat{L}_{\Dtrr}$ (eq.~\ref{eq:ml-forgetting}) using $\tau$ steps of SGD with mini-batch size $B$. Let $\gamma = 1-\mu^2/\beta^2$, where $\beta$ is the smoothness constant of the loss in $\cref{eq:private-training-linear}$. Consider a sequence of $K$ equally sized forgetting requests $\{\Dtrff^{1},\Dtrff^{2},\ldots,\Dtrff^{K}\}$ and let $\w_u^{K}$ be the weights obtained after the $K$ requests using \cref{eq:ml-forgetting}. Then we have the following bound on the amount of information remaining in the weights $\w_{u}^{K}$ about $\cup_{k=1}^{K} \Dtrff^{k}$
{\normalfont
\begin{align}
\label{eq:forgetting-bound}
\mathcal{I} (\cup_{k=1}^{K} \Dtrff^k) \leq \dfrac{\!\!\!\!\!\! \aoverbrace[L1R]{\vphantom{\Big(} \gamma^{\tau}}^{\substack{\text{forgetting}\\\text{ steps}}} \!\!\! c_0 \Big(\!\!\!\aoverbrace[L1R]{\vphantom{\Big(}c_1 \dfrac{r^2}{\sigma^2}}^{\substack{\text{ratio to}\\\text{forget}}} \!\! + \!\! \aoverbrace[L1R]{\vphantom{\Big(}d}^{\substack{\text{num.}\\ \text{params}}} \!\!\!\Big)  +  \aoverbrace[L1R]{\dfrac{c_2}{B\sigma^2}}^{\substack{\text{batch}\\ \text{size}}}}{1 - (1 + \alpha) \gamma^{\tau}}.
\end{align}
}
where $c_0, c_1, c_2 > 0$, $r = |\Dtrff^{k}|/|\D|$ and  $0 < \alpha < 1/\gamma^{\tau}-1$, $d=\operatorname{dim}(\w)$ and $\gamma < 1$.
\end{theorem}

In \cite{neel2020descenttodelete} a similar probabilistic bound is given on the distance of the scrubbed weights from the optimal weights for strongly convex Lipschitz loss functions trained using projected GD. We prove our bound for the more general case of a convex loss function with $L_2$ regularization trained using SGD (instead of GD) and also bound the remaining information in the weights.

\textbf{Role of $\sigma$.} We make some observations regarding \cref{eq:forgetting-bound}. First, increasing the variance $\sigma^2$ of the noise added to the weights after the forgetting step further reduces the possible leakage of information from an imperfect approximation. Of course, the downside is that increasing the noise may reduce the performance of the model (see \Cref{fig:info_vs_noise_epochs} (top) for the trade-off between the two).

\textbf{Forgetting with more iterations.} Running the algorithm $\A$ for an increasing number of steps $\tau$ improves the accuracy of the forgetting step, and hence reduces the amount of remaining information. We confirm this empirically in \Cref{fig:info_vs_noise_epochs} (bottom). Note however that there is a diminishing return. This is due to the variance of the stochastic optimization overshadowing gains in accuracy from longer optimization (see the additive term depending on the batch size). Increasing the batch-size, $B$ in \cref{eq:ml-forgetting} reduces the variance of the estimation and leads to better convergence.

\textbf{Fraction of data to forget.} Finally, forgetting a smaller fraction $r=|\Dtrff^k|/|\D|$ of the data is easier. On the other hand, increasing the number of parameters $d$ of the model may make the forgetting more difficult.

\begin{figure}[t]
    \centering
    \includegraphics[width=1.0\linewidth]{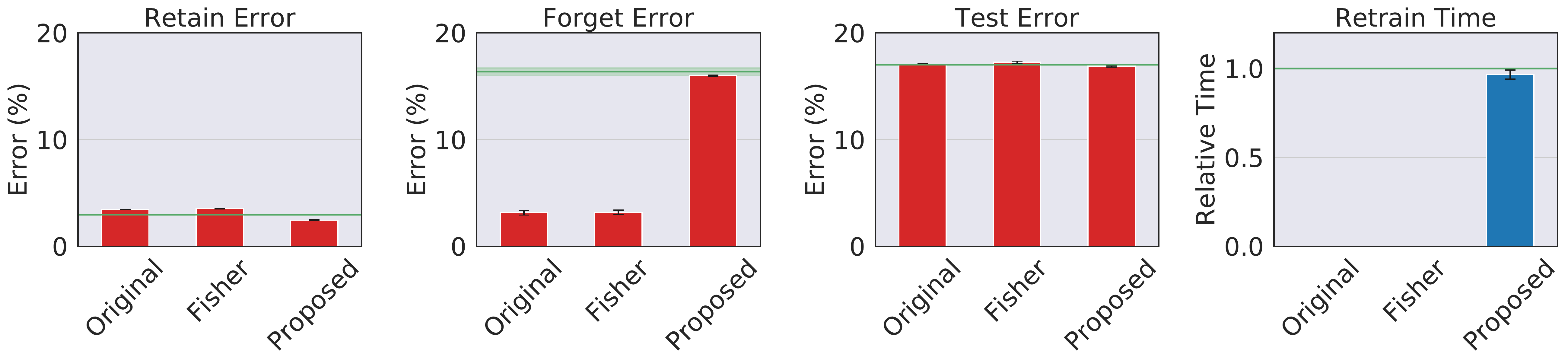}
    \includegraphics[width=1.0\linewidth]{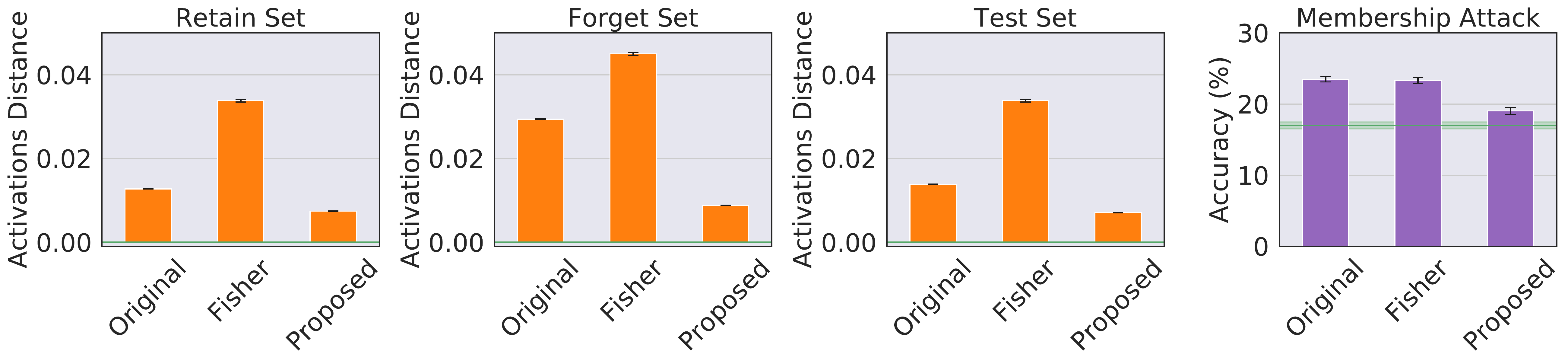}
    \caption{
    \textbf{Read-out functions for different forgetting methods.} We forget a subset of 10\% of the training data through 10 equally-size sequential deletion requests using different forgetting methods, and show the value of the several readout functions for the resulting scrubbed models. Ideally, the value of the readout function should be the same as the value (denoted with the green area) obtained on a model retrained  from scratch without those samples. Closer to the green area is better. \textbf{(Original)} denotes the trivial baseline where we do apply any forgetting procedure. 
    \textbf{(Fisher)} Adds Fisher noise as as described in (\cite{golatkar2019eternal}), \textbf{(\method)} The model obtained with our method after forgetting.
    In all cases, we observe that \method obtains a model that is indistinguishable from one trained from scratch without the data, whereas the other methods fail to do so. This is particularly the case for the \textit{Retrain Time} readout functions, which exploits full knowledge of the weights and it is therefore more difficult to defend against.
    }\vspace{-0.25cm}
    \label{fig:readout-main}
\end{figure}

\vspace{-0.25cm}
\section{Experiments}
\label{sec:experiments}

We use a ResNet-50 \cite{he2016deep} as the model $f_{\w}(\x)$ in ML-Forgetting.
Unless specified otherwise, we forget $10\%$ of randomly chosen training data in all the experiments through 10 sequential forgetting requests each of size $1\%$. In the appendix, we also provide results for forgetting an entire class and show that our method is invariant to the choice of the subset to be forgotten. More experimental details can be found in the appendix.

\textbf{Datasets used.} We test our method on the following image classification tasks: Caltech-256~\cite{griffin2007caltech}, MIT-67~\cite{sharif2014cnn}, Stanford Dogs~\cite{khosla2011novel}, CUB-200~\cite{wah2011caltech}, FGVC Aircrafts~\cite{maji2013fine}, CIFAR-10~\cite{krizhevsky2009learning}. Readout function and forgetting-accuracy trade-off plots for MIT-67, StanfordDogs, CUB-200 and CIFAR-10 can be found in the appendix.

\subsection{Readout functions}
\label{sec:readout-functions}
The forgetting procedure should be such that an attacker with access to the scrubbed weights $\w$ should not be able to construct some function $R(\w): \R^{d} \rightarrow \R$, which will leak information about the set to forget $\Dtrff$. More precisely the scrubbing procedure should be such that for all $R(\w)$:
\begin{equation}
    \KL{\mathbf{P}(\underbrace{R(S(\w, \D, \Dtrff))|\D}_{\substack{\text{readout on weights} \\ \text{after forgetting $\Dtrff$}}})}{\mathbf{P}(\underbrace{R(S_0(\w))|\Dtrr}_{\substack{\text{readout on weights} \\ \text{after re-training on $\Dtrr$}}}} = 0
    \label{equation:readout-information}
\end{equation}
where $S_0(\w)$ is some baseline function that does not depend on $\Dtrff$ (it only depends on the subset to retain $\Dtrr$ = $\D - \Dtrff$). Here $\mathbf{P}(\w|\D)$, $\mathbf{P}(\w|\Dtrr)$ corresponds to the distribution of weights (due to the stochastic training algorithm) obtained after minimizing the empirical risk on $\D$, $\Dtrr$ respectively. $S(\w, \D, \Dtrff)$ corresponds to the scrubbing update defined in \cref{eq:ml-forgetting}. We choose $S_0(\w) = \w + z$, where $z \sim \mathcal{N}(0, \sigma^2 I)$. For an ideal forgetting procedure, the value of the readout functions (or evaluation metrics) should be same for a model obtained after forgetting $\Dtrff$ and re-trained from scratch without using $\Dtrff$.  
Some common choice of readout functions include (see \Cref{fig:readout-main}):
\begin{enumerate}[wide, labelindent=0pt]
    \item \textbf{Error on $\Dtrr$, $\Dtrff$, $\Dte$:} The scrubbed and the re-trained model (from scratch on $\Dtrr$) should have similar accuracy on all the three subsets of the data
    \item \textbf{Re-learn Time:} We fine-tune the scrubbed (model after forgetting) and re-trained model for a few iterations on a subset of the training data (which includes $\Dtrff$) and compute the number of iterations it takes for the models to re-learn $\Dtrff$. An ideal forgetting procedure should be such that the re-learn time should be comparable to the re-trained model (we plot the relative re-train time in \Cref{fig:readout-main}). Re-learn time serves a proxy for the amount of information remaining in the weights about $\Dtrff$ (see \Cref{fig:readout-main}).
    \item \textbf{Activation Distance:} We compute the distance between the final activations of the scrubbed weights and the re-trained model ($\w_{\Dtrr}$) on different subsets of data. More precisely we compute the following: $\E_{\x \in \D'}[\|\softmax(f_{{\w}}(\x)) - \softmax(f_{{\w}_{\Dtrr}}(\x))\|_{1}]$, where $\D'=\Dtrr,\Dtrff,\Dte$. We compare different $\w$ corresponding to the original weights without any forgetting, weights after adding Fisher noise and ML-forgetting (see \Cref{fig:readout-main}). This serves as a proxy for the amount of information remaining in the activations about $\Dtrff$.
    \item \textbf{Membership Attack:} We construct a simple yet effective membership attack similar to \cite{golatkar2020forgetting} using the entropy of the model output. Ideally, a forgetting procedure should have the same attack success as a re-trained model (which is what we observe, see \Cref{fig:readout-main}).
\end{enumerate}

\subsection{Complete vs Stochastic residual gradient}
\begin{figure}[t]
    \centering
    \includegraphics[width=1.0\linewidth]{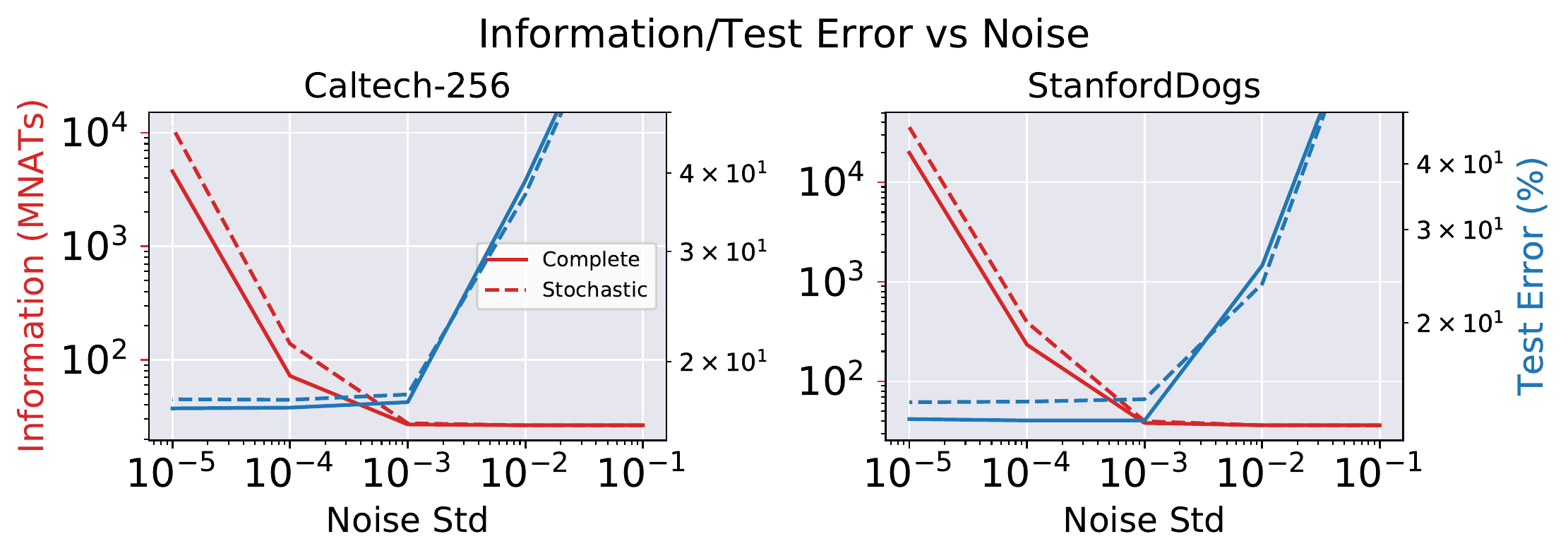}
    \caption{\textbf{Comparison of complete and stochastic residual gradient estimation for forgetting.} \method uses the complete residual gradient for the forgetting step, exploiting the fact that the loss function is quadratic. However, one can also estimate it stochastically, which is equivalent to fine-tuning on the remaining data to forget. Here we show that indeed both methods work but -- when using the same number of steps -- complete estimate gives a better solution due to smaller variance and faster convergence (lower test error and information leakage).
    }\vspace{-1em}
    \label{fig:info_vs_noise_compare_with_ft}
\end{figure}
In \cref{eq:ml-forgetting} we compute the residual gradient $g_{\Dtrr}(\w_{u})$ completely once over the remaining data instead of estimating that term stochastically using $\A$. In \Cref{fig:info_vs_noise_compare_with_ft}, we compare both the methods of computing the residual gradient. We show that in the ideal region of noise (i.e. $\sigma \in [10^{-5},10^{-3}]$), both the remaining information and test error after forgetting (10\% of the data through 10 requests) is lower when computing the residual gradient completely.

\subsection{Effect of choosing different core datasets}
\begin{figure}[t]
    \centering
    \includegraphics[width=0.95\linewidth]{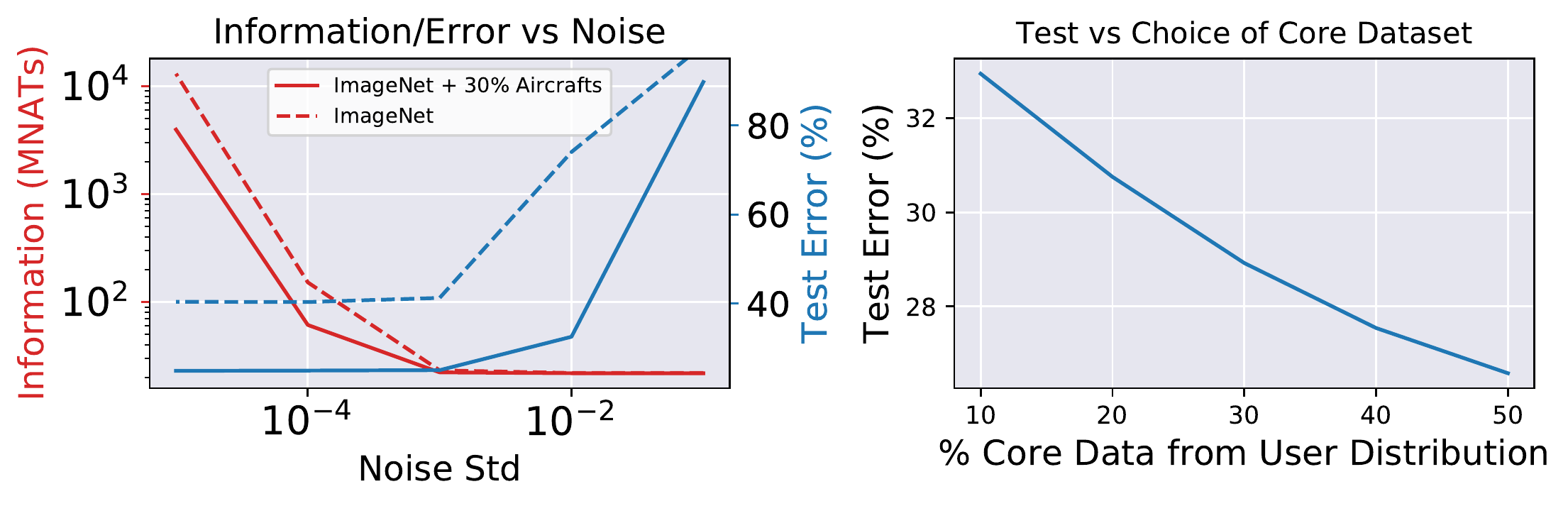}
    \caption{\textbf{Effect of using a core data close to the task.} Plot of the remaining information and test error on \textit{Aircrafts} using (a) generic ImageNet core data, and (b) ImageNet pre-training + 30\% of the \textit{Aircrafts}. When the core data contain information close to the user task that the network does not need to forget, \method can exploit this to create better core-weights and a correspondingly better linearized model. This improves both the accuracy of the model and makes forgetting easier, as seen from the accuracy-forgetting curves in the plot.
    }
    \label{fig:info_vs_noise_core_data_effect_aircrafts}
\end{figure}
For fine-grained datasets like FGVC-Aircrafts and CUB-200, we show that if the core data has some information about the user task, then it improves forgetting significantly both in terms of the remaining information and the test accuracy. In \Cref{fig:info_vs_noise_core_data_effect_aircrafts}, we show that using ImageNet + 30\% of the Aircrafts (we assume that we are not asked to forget this 30\% of the data) as core data and 100\% of the Aircrafts as the user data, performs much better than simply using ImageNet as core. In \Cref{fig:info_vs_noise_core_data_effect_aircrafts} (right), we also show that increasing the percentage of user distribution in the core data improves the test accuracy of the Mixed-Linear model.

\vspace{-0.1cm}
\section{Conclusion}

We provide a practical forgetting procedure to remove the influence of a subset of the data from a trained image classification model. We achieve this by linearizing the model using a mixed-privacy setting which enables us to split the weights into a set of core and forgettable user weights. When asked to delete all the user data, we can simply discard the user weights. The quadratic nature of the training loss enables us to efficiently forget a subset of the user data without compromising the accuracy of the model. In terms of the time-complexity, we only need 2-3 passes over the dataset per forgetting query for removing information from the weights rather than the 50 re-training epochs, thus, providing a 16$\times$ or more speed-up per request (see \Cref{fig:info_vs_noise_epochs}). We test the forgetting procedure against various read-out functions, and show that it performs comparably to a model re-trained from scratch (the ideal paragon). Finally, we also provide theoretical guarantees on the amount of remaining information in the weights and verify the behavior of the information bounds empirically through extensive evaluation in \Cref{fig:info_vs_noise_epochs}.

Our forgetting procedure heavily relies on the strongly convex nature of the loss landscape which is induced by $L_2$ regularization (increasing it improves forgetting but compromises accuracy). The quality of forgetting also relies on the subset of data to be forgotten, however, we will leave this for the future work. Even though we provide a forgetting procedure for deep networks by linearizing them without compromising their accuracy, directly removing information from highly non-convex deep networks efficiently still remains an unsolved problem at large.

\clearpage

\newpage

{\small
\bibliographystyle{ieee_fullname}
\bibliography{egbib}
}

\clearpage
\newpage

\appendix
\section*{Appendix}
In this appendix we provide additional experiments (\Cref{app:additional-exp}), experimental details (\Cref{app:experimental-details}) and theoretical results (\Cref{app:theoretical-results}).

\section{Additional Experiments}
\label{app:additional-exp}
\subsection{Forgetting an entire class}
\begin{figure}[H]
    \centering
    \includegraphics[width=1.0\linewidth]{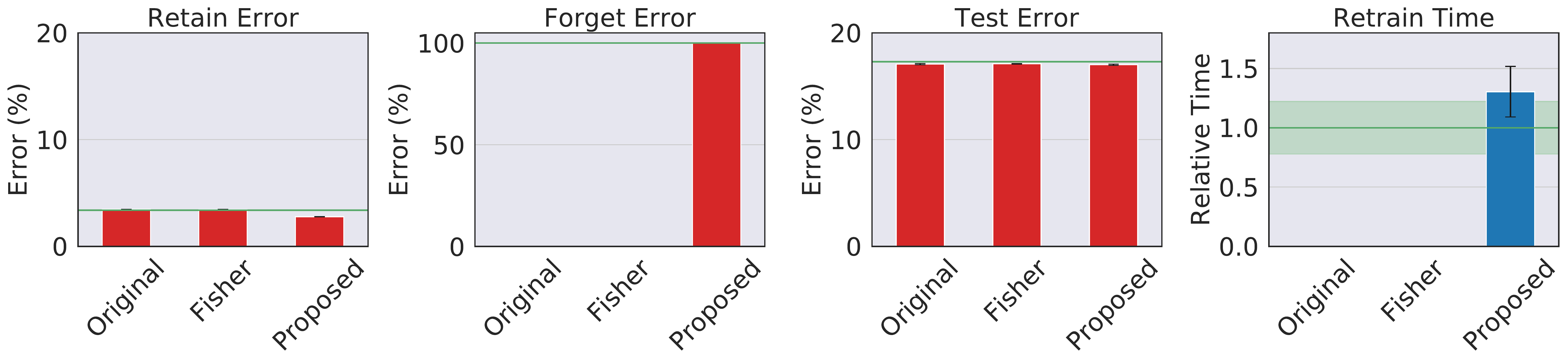}
    \includegraphics[width=1.0\linewidth]{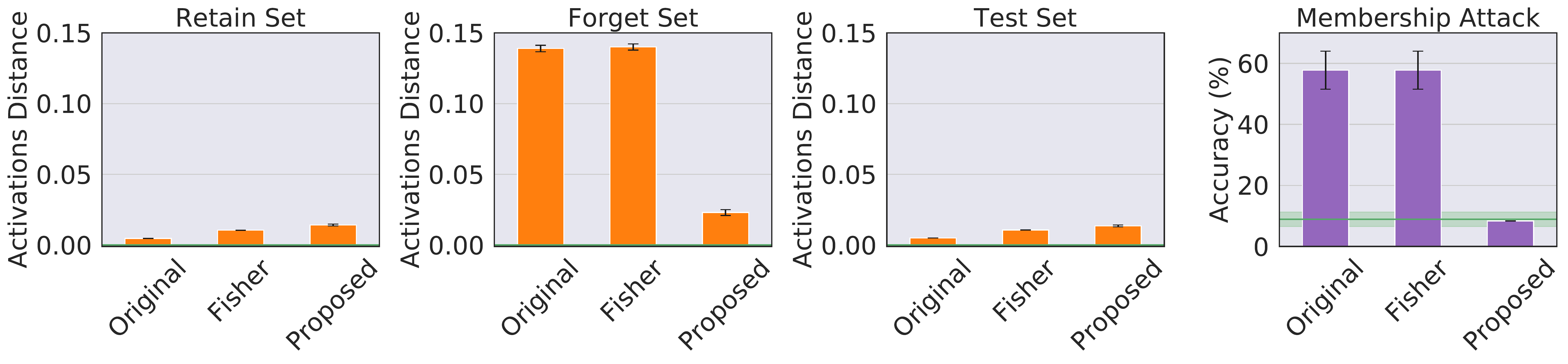}
    \caption{Readout function plot similar to \Cref{fig:readout-main} for Caltech-256 dataset, where we forget an entire class rather a sequence of randomly sampled data subsets.\vspace{-0.5cm}}
    \label{fig:readout-caltech-class}
\end{figure}

\begin{figure}[H]
    \centering
    \includegraphics[width=1.0\linewidth]{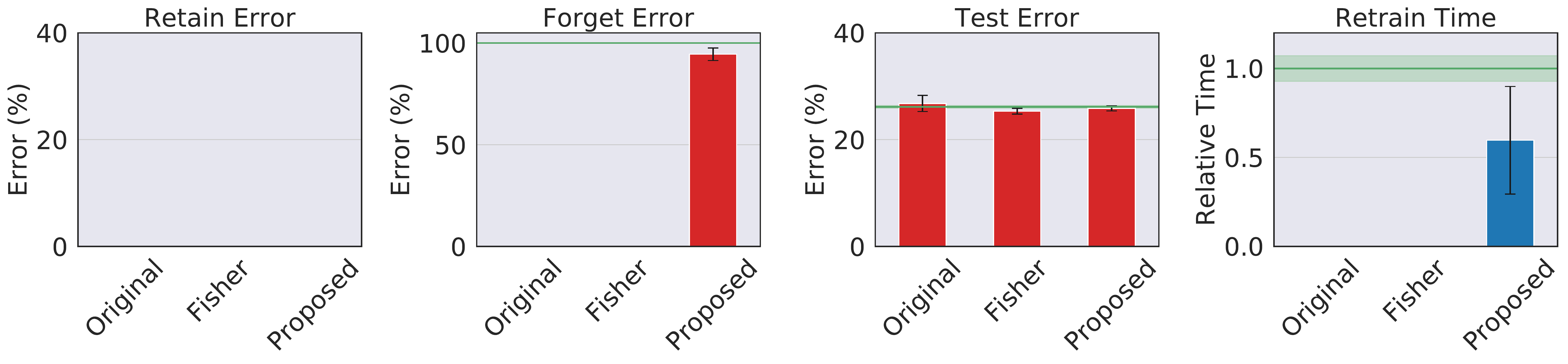}
    \includegraphics[width=1.0\linewidth]{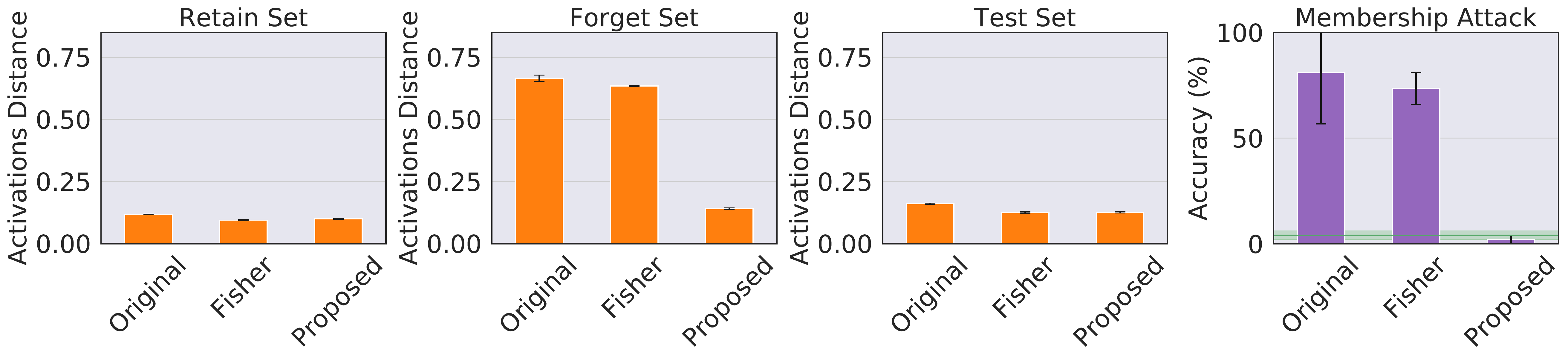}
    \caption{Readout function plot similar to \Cref{fig:readout-main} for FGVC-Aicrafts dataset, where we forget an entire class rather a sequence of randomly sampled data subsets.}
    \label{fig:readout-fgvc-aircrafts-class}
\end{figure}

In the main paper we considered forgetting a random subset of 10\% of the training data. Here we consider instead the problem of completely forgetting all samples of a given class in a single forgetting request. In \Cref{fig:readout-caltech-class,fig:readout-fgvc-aircrafts-class}, we observe that also in this setting our proposed method outperforms other methods and is robust to different readout functions. Note that for the case of removing an entire class the target forget error (i.e. the error on the class to forget) is 100\%.

\subsection{Role of $L_2$-Regularization}
\begin{figure}[H]
    \centering
    \includegraphics[width=1.0\linewidth]{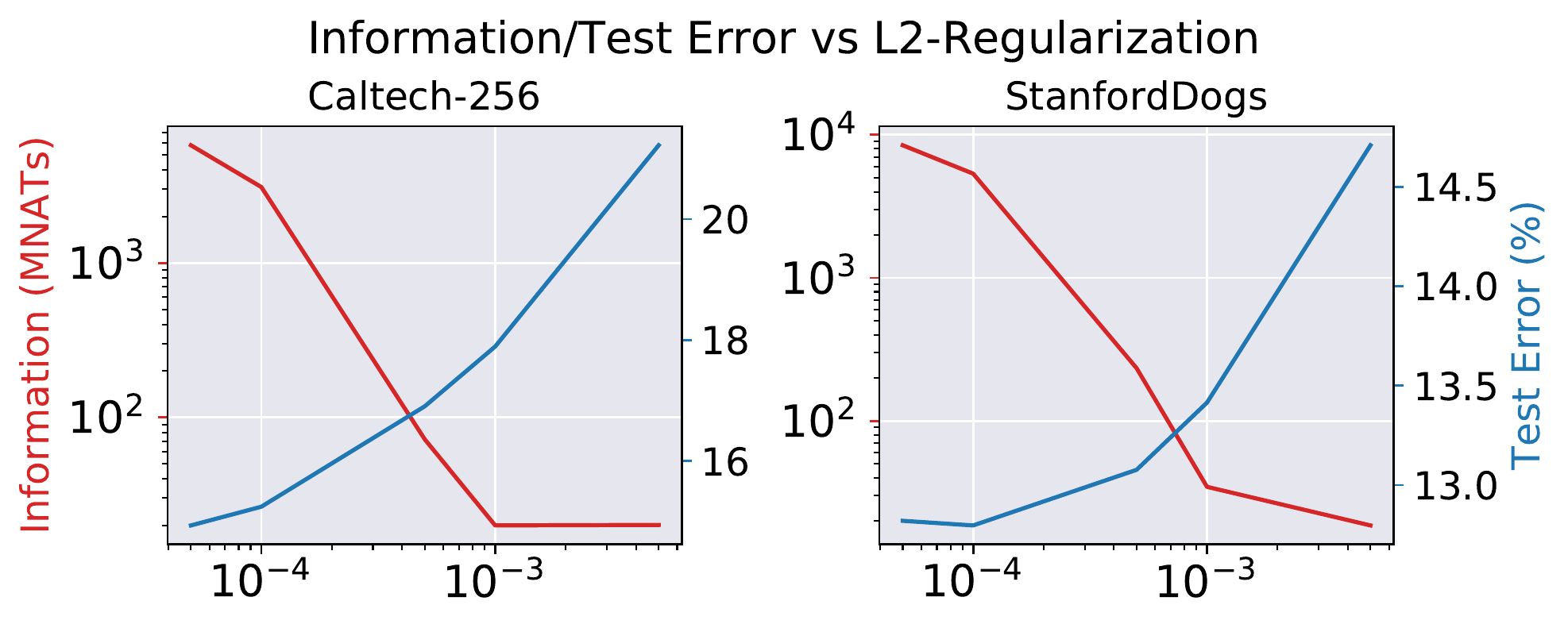}
    \caption{Plot of the amount of remaining information and test error vs the $L_2$ regularization coefficient. We forget 10\% of the training data sequentiall through 10 forgetting request.}
    \label{fig:info_test_vs_l2}
\end{figure}

We plot the amount of remaining information and the test error as a function of the $L_2$ regularization coefficient. Note that instead of incorporating weight decay directly in the optimization step, as it is often done, we explicitly add the $L_2$ regularization to the loss function. As expected theoretically (\Cref{theorem-information-formal}), increasing the regularization coefficient makes the training optimization problem more strongly convex, which in turn makes forgetting easy. However, increasing weight decay too much also hurts the accuracy of the model. Hence there is a trade-off between the amount of remaining information and the amount of regularization with respect to the regularization. We plot the trade-off in \Cref{fig:info_test_vs_l2}.

\subsection{More experiments using SGD for forgetting}
We repeat the same experiments as in \Cref{fig:readout-main} on the following datasets: \textit{Stanford Dogs, MIT-67, CIFAR-10, CUB-200, FGVC Aircrafts}. Overall, we observe consistent results over all datasets.

\begin{figure}[H]
    \centering
    \includegraphics[width=1.0\linewidth]{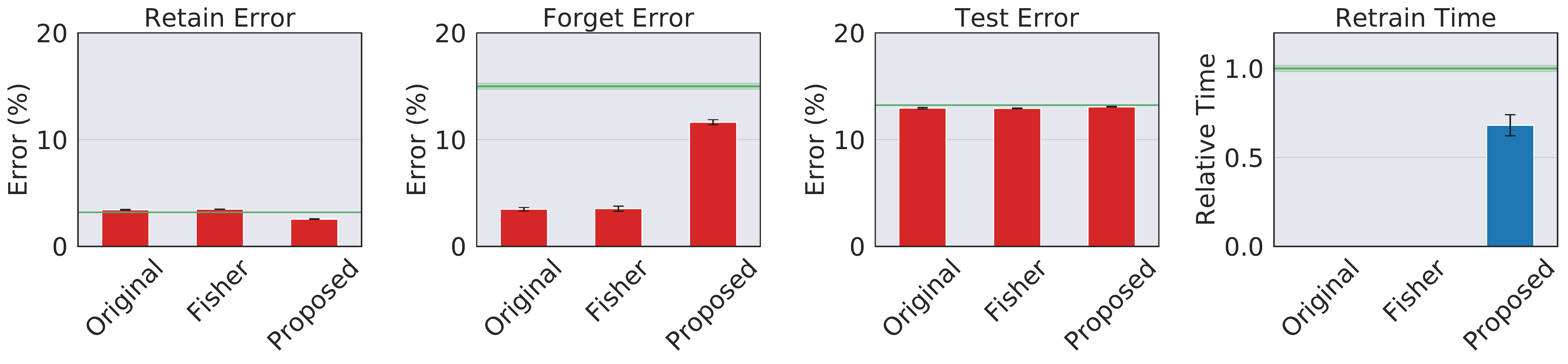}
    \includegraphics[width=1.0\linewidth]{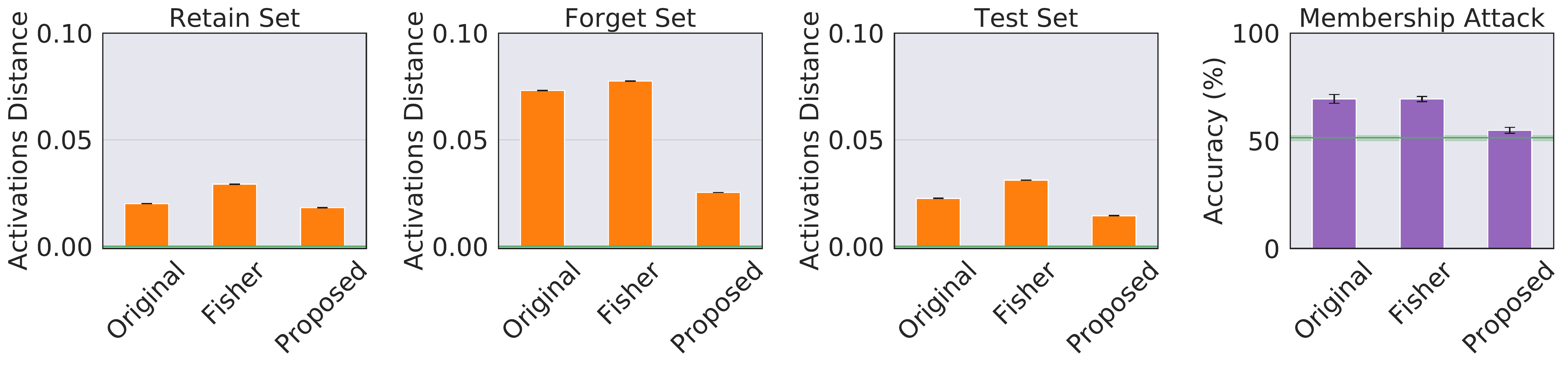}
    \caption{Same experiments as \Cref{fig:readout-main} for StanfordDogs.}
    \label{fig:readout-stanforddogs}
\end{figure}
\vspace{-0.25cm}
\begin{figure}[H]
    \centering
    \includegraphics[width=1.0\linewidth]{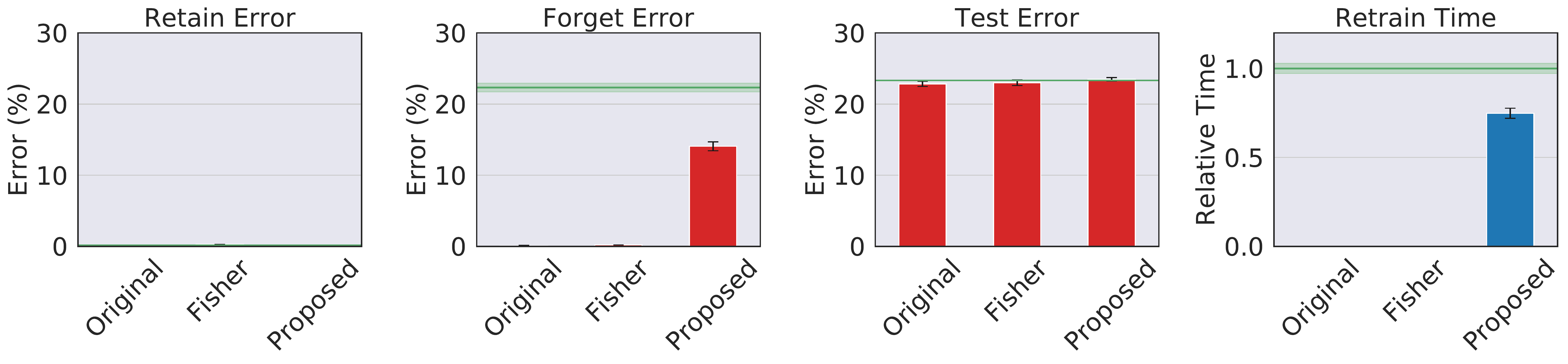}
    \includegraphics[width=1.0\linewidth]{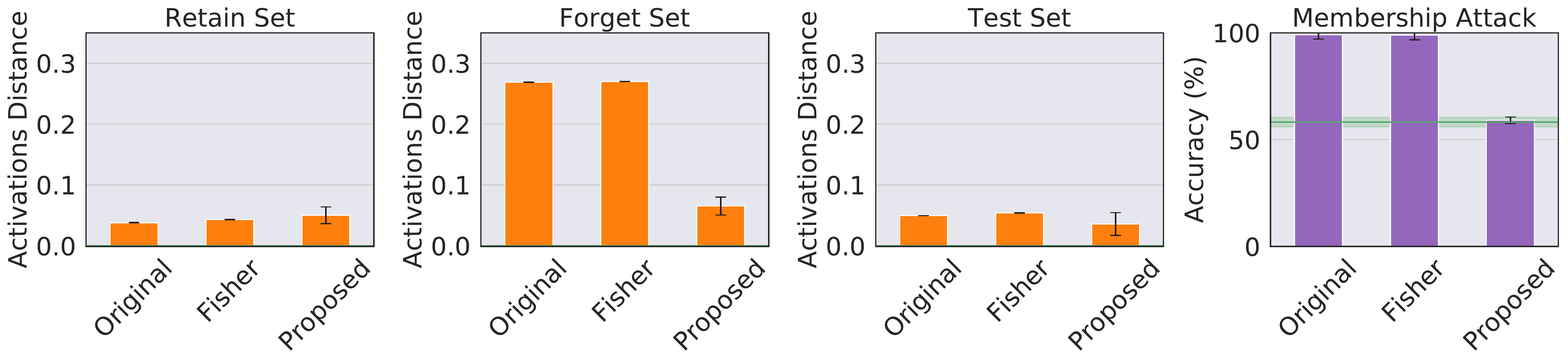}
    \caption{Same experiments as \Cref{fig:readout-main} for MIT-67.}
    \label{fig:readout-mit67}
\end{figure}
\vspace{-0.5cm}
\begin{figure}[H]
    \centering
    \includegraphics[width=1.0\linewidth]{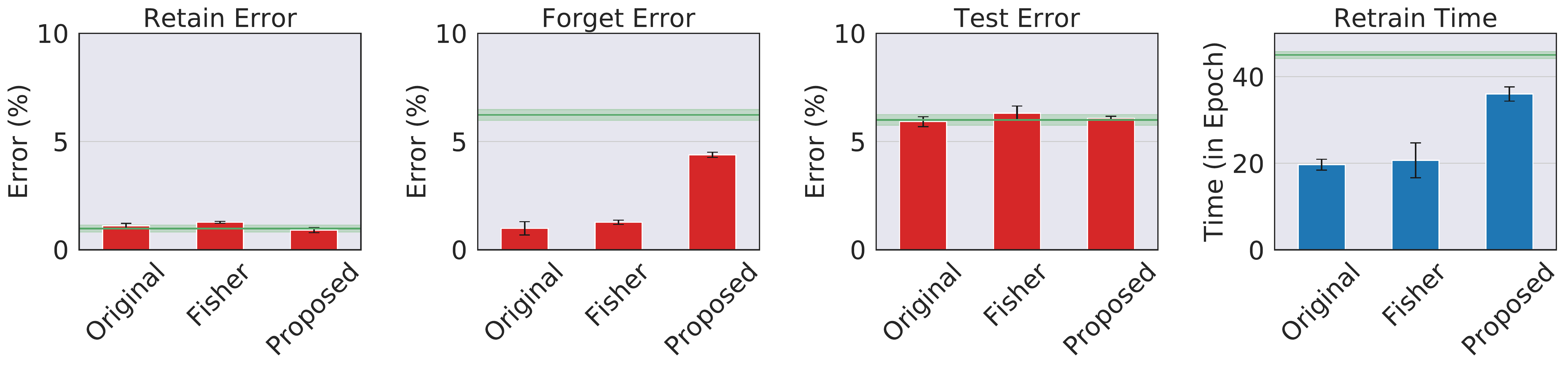}
    \includegraphics[width=1.0\linewidth]{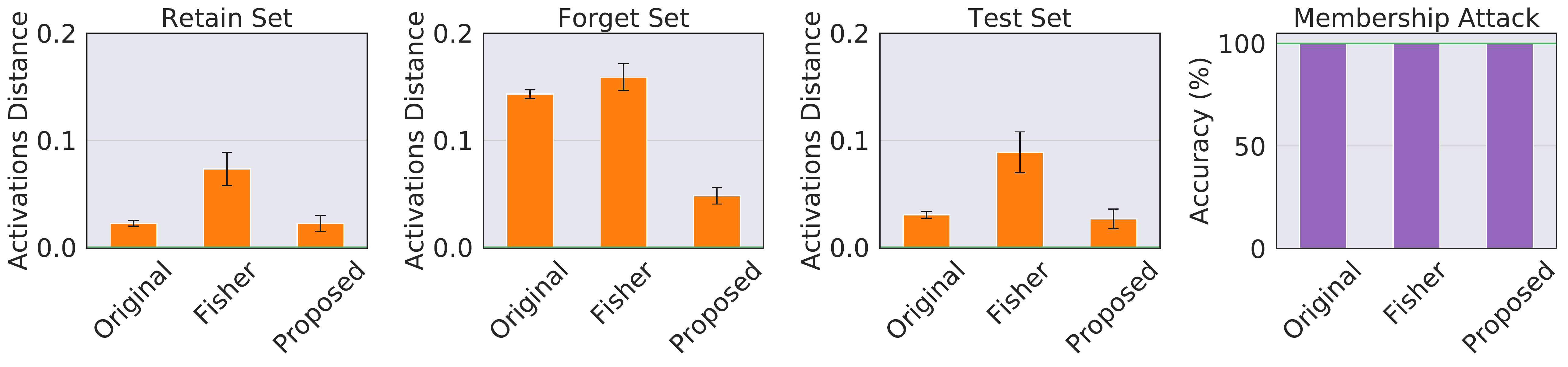}
    \caption{Same experiments as \Cref{fig:readout-main} for CIFAR-10.}
    \label{fig:readout-cifar10}
\end{figure}
\vspace{-0.75cm}
\begin{figure}[H]
    \centering
    \includegraphics[width=1.0\linewidth]{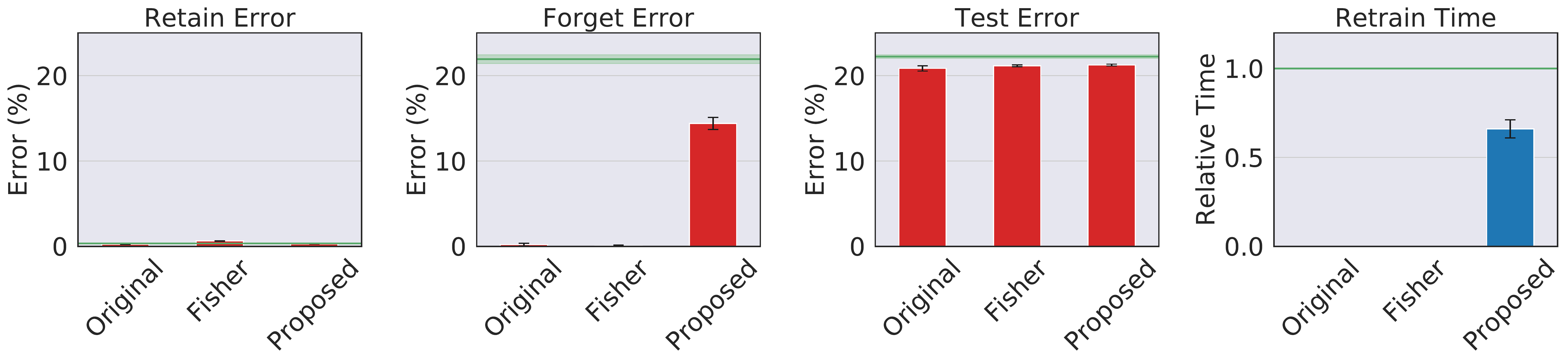}
    \includegraphics[width=1.0\linewidth]{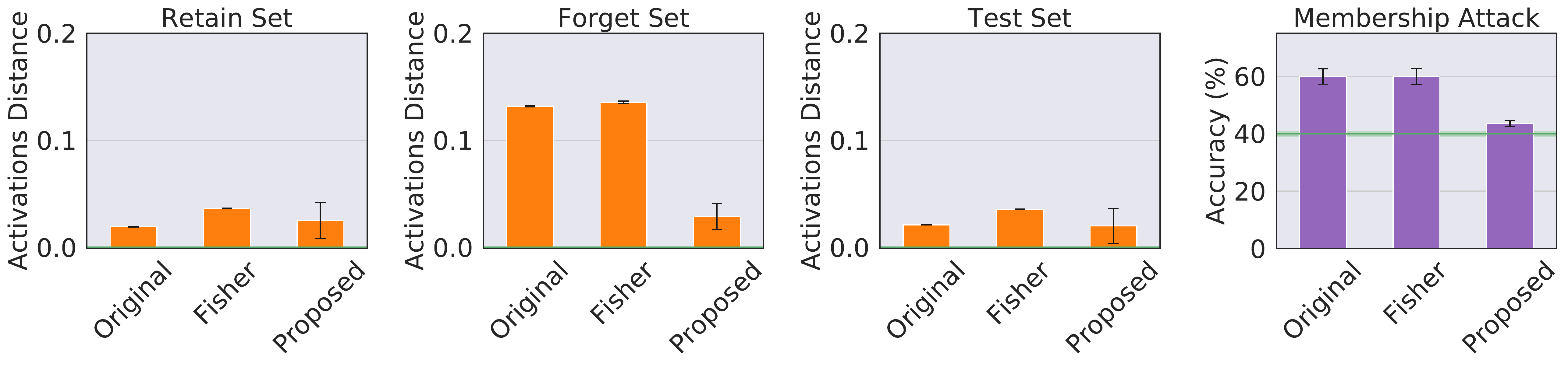}
    \caption{Same experiments as \Cref{fig:readout-main} for CUB-200.}
    \label{fig:readout-cub200}
\end{figure}
\vspace{-0.75cm}
\begin{figure}[H]
    \centering
    \includegraphics[width=1.0\linewidth]{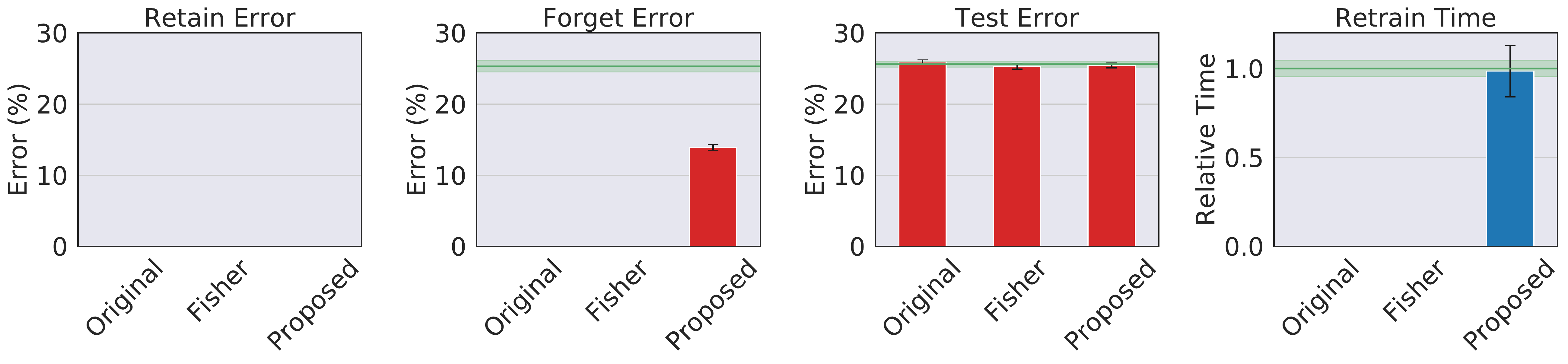}
    \includegraphics[width=1.0\linewidth]{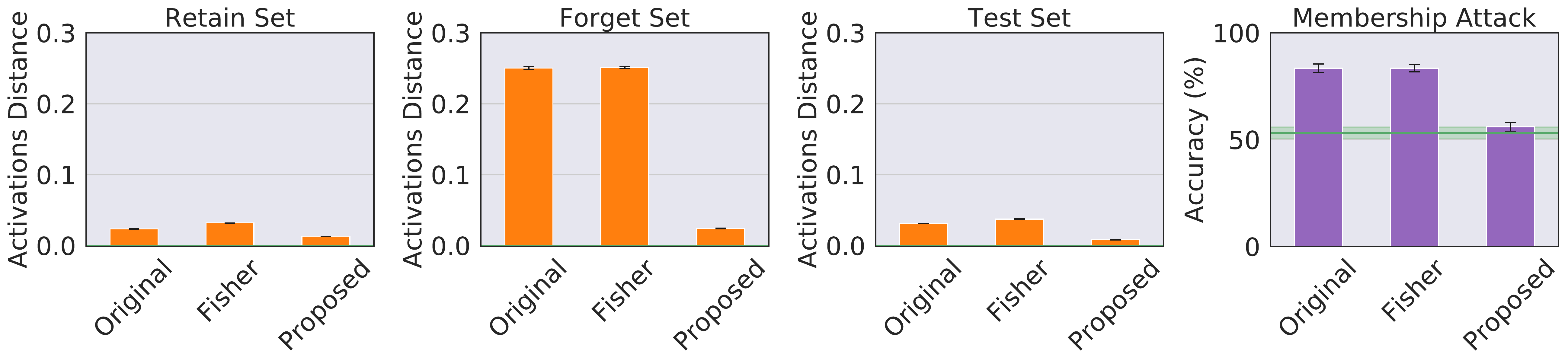}
    \caption{Same experiments as \Cref{fig:readout-main} for FGVC-Aircrafts.\vspace{-0.5cm}}
    \label{fig:readout-fgvc_aircrafts}
\end{figure}
\subsection{Information vs Noise/Epochs}
\begin{figure}[H]
    \centering
    \includegraphics[width=1.0\linewidth]{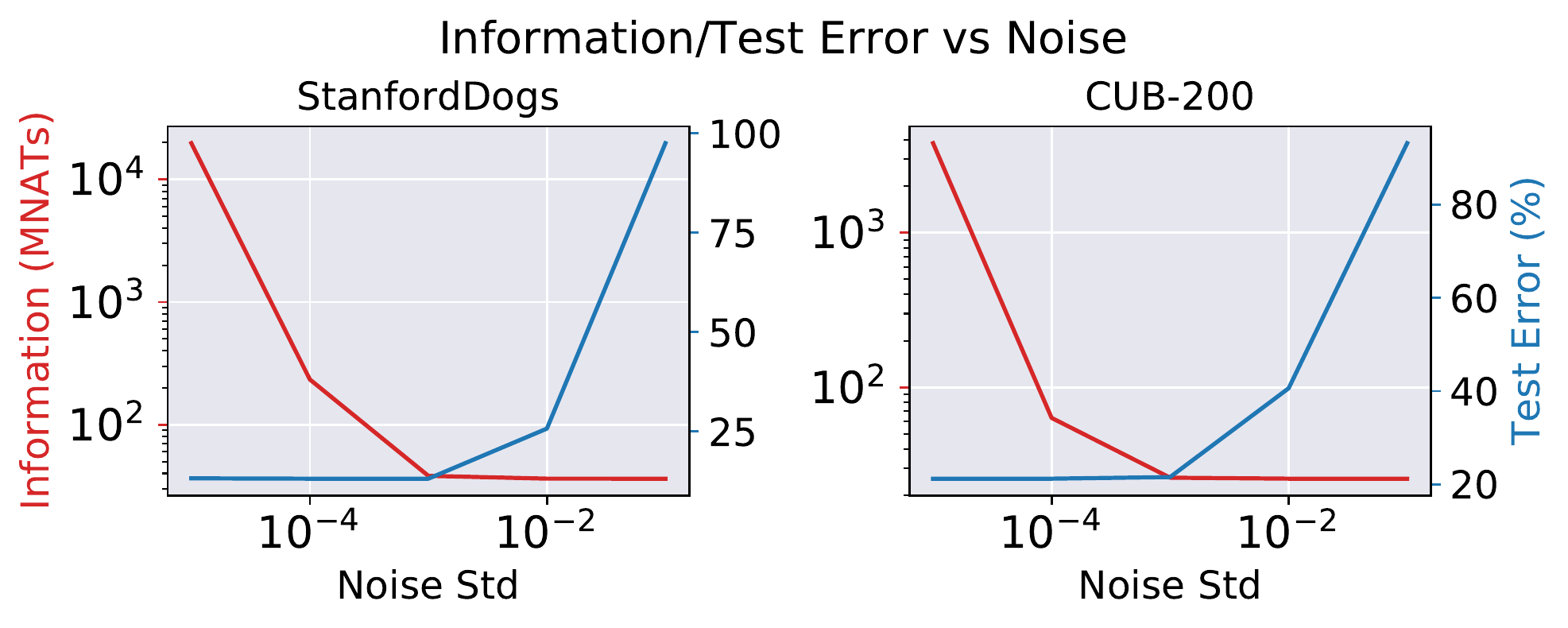}
    \includegraphics[width=1.0\linewidth]{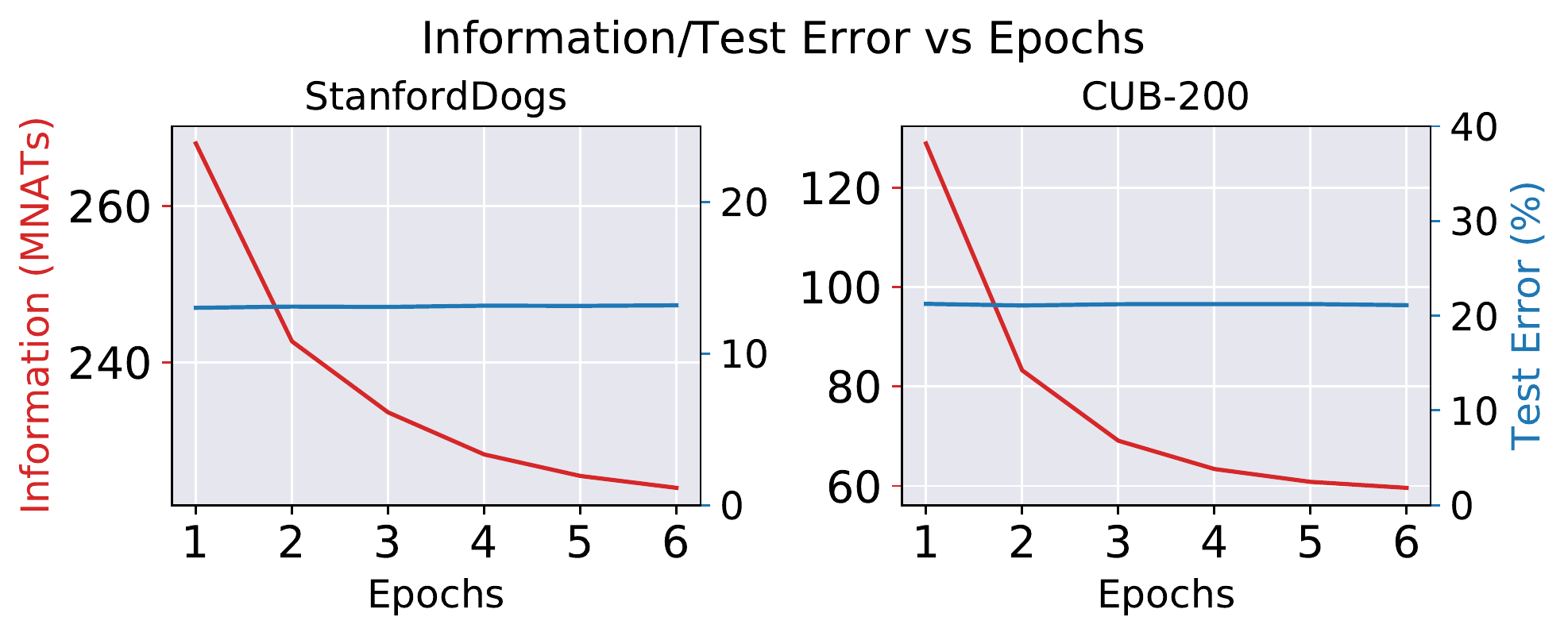}
    \caption{Same experiment as \Cref{fig:info_vs_noise_epochs} for Stanforddogs and CUB-200 datasets.}
    \label{fig:info_vs_noise_epochs_stanforddogs_cub}
\end{figure}
\vspace{-0.25cm}
\begin{figure}[H]
    \centering
    \includegraphics[width=1.0\linewidth]{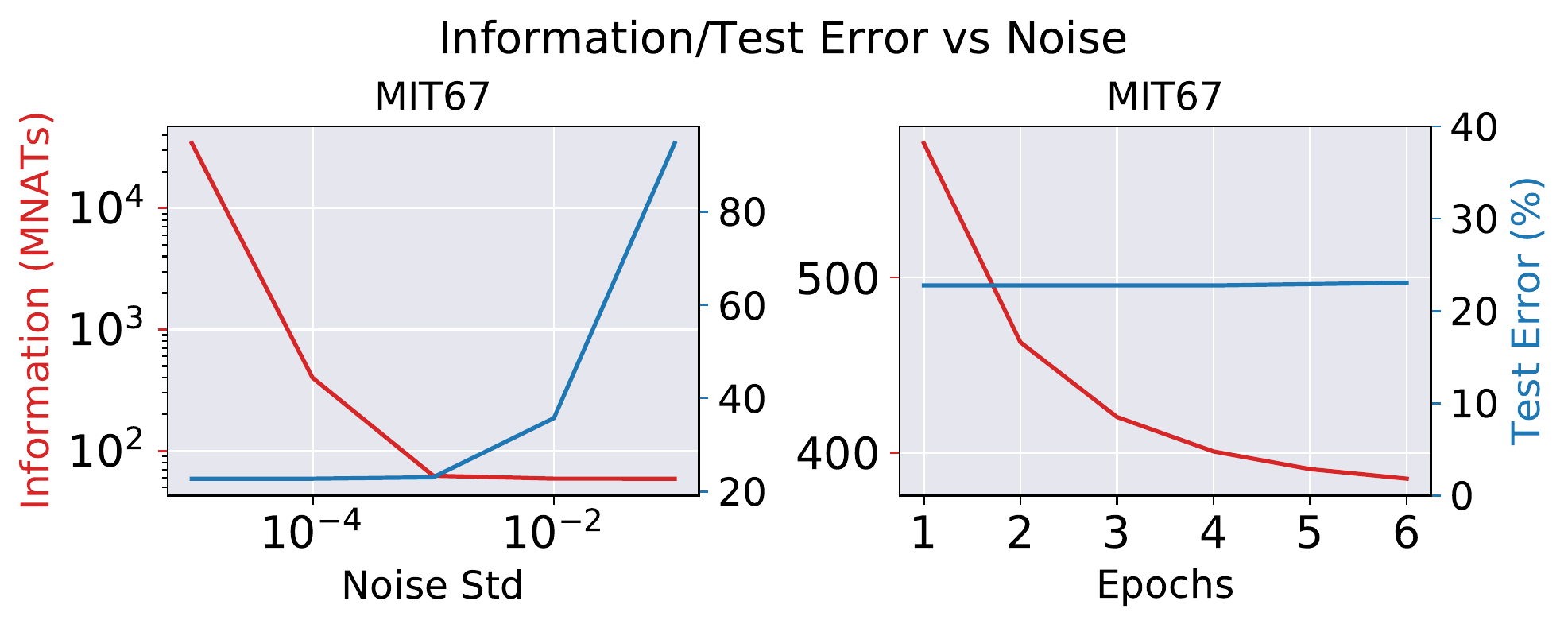}
    \caption{Same experiment as \Cref{fig:info_vs_noise_epochs} for MIT67.}
    \label{fig:info_vs_noise_epochs_stanforddogs_cub}
\end{figure}

\section{Experimental Details}
\label{app:experimental-details}
We use a ResNet-50 pre-trained on ImageNet. For the plots in \Cref{fig:error-comparison}, we train \method model using SGD for 50 epochs with batch size 64, learning rate lr=0.05, momentum=0.9, weight decay=0.00001 where the learning rate is annealed by 0.1 at 25 and 40 epochs. We explicitly add the $L_2$ regularization to the loss function instead of incorporating it in the SGD update equation. We only linearize the final layers of ResNet-50 and scale the one-hot vectors by 5 while using the MSE loss. For fine-grained datasets, FGVC-Aircrafts and CUB-200, in addition to the ImageNet pre-training, we also pre-train them using randomly sampled 30\% of the training data (which we assume is part of the core set). 

For the training the \method model in the readout functions and information plots using SGD, we use the same experimental setting as above with a increased weight decay=0.0005 for Caltech-256,StanfordDogs and CIFAR-10 and 0.001 for MIT-67,CUB-200 and FGVC-Aircrafts. We use a higher value of weight decay to increase the strong convexity constant of the training loss function, which facilitates forgetting (see \Cref{lemma-sensitivity}). 

For forgetting using \method model in the readout function/information plots using SGD (\method to minimize \cref{eq:conjugate-problem}), we use momentum=0.999 and decrease the learning rate by 0.5 per epoch. We run SGD for 3 epochs with an initial lr=0.01 for Caltech-256, StanfordDogs and CIFAR-10 and run it for 4 epochs with initial lr=0.025 for MIT-67, CUB-200 and FGVC-Aircrafts. 
\clearpage
\onecolumn

\section{Theoretical Results}
\label{app:theoretical-results}
\begin{lemma}
Let $\x,\y$ be two random vectors such that $\E \|\x\|^2, \E \|\y\|^2 > 0$. Then we have the following, for any $\alpha > 0$:
\[\E\|\x + \y\|^2 \leq (1+\alpha)\E \|\x\|^2 + (1+\dfrac{1}{\alpha})\E \|\y\|^2\]
\label{lemma-young}
\end{lemma}
\begin{proof}
\begin{align}
\nonumber \E\|\x + \y\|^2 &= \E(\|\x\|^2 + \|\y\|^2 + 2\big<\x,\y\big>)\\
\nonumber &\leq \E(\|\x\|^2 + \|\y\|^2 + 2|\big<\x,\y\big>|)\\
\nonumber &\stackrel{(a)}{\leq} \E \bigg(\|\x\|^2 + \|\y\|^2 + 2\sqrt{\|\x\|^2}\sqrt{\|\y\|^2} \bigg)\\
\nonumber &= \E \bigg(\|\x\|^2 + \|\y\|^2 + 2\sqrt{\|\x\|^2 \alpha}\sqrt{\dfrac{\|\y\|^2}{\alpha}} \bigg)\\
\nonumber &\stackrel{(b)}{\leq} \E \bigg(\|\x\|^2 + \|\y\|^2 + \|\x\|^2 \alpha + \|\y\|^2 \dfrac{1}{\alpha} \bigg)\\
&= (1+\alpha)\E \|\x\|^2 + (1+\dfrac{1}{\alpha})\E \|\y\|^2 \label{equation:young}
\end{align}

for any $\alpha > 0$, where (a) follows from the Cauchy-Schwarz inequality and (b) follows from the AM-GM inequality. 
\end{proof}

\begin{lemma}
Let $C = \{\w | \w \in \R^d \text{ and } \|\w\| \leq R\ < \infty \}$ and $\ell(\w): \R^d \rightarrow \R^+$ be a strongly convex function with $\text{max}_{\w \in C} \ell(\w) < \infty$, $G \triangleq \text{max}_{\w \in C} \|\nabla \ell(\w)\| < \infty $ and $\w^* = \text{argmin}_{\w \in \R^d} \ell(\w)$ such that $\|\w^*\| < \infty$. Then $\forall \w_1, \w_2 \in C$, we have that $|\ell(\w_1) - \ell(\w_2)| \leq G \|\w_1 - \w_2\|$. When $\ell(\w)$ is also quadratic with $ \beta = \lambda_{\text{Max}}(\nabla^2\ell(\w))$, the maximum eigen value of the Hessian, we have that $G=\beta \big(R + \|\w^*\| \big)$.
\label{lemma:lipschitz}
\end{lemma}

\begin{proof}
Let $g(t) = \ell(\w_1 t + \w_2 (1-t))$, where $t \in [0,1]$ then from Mean Value Theorem (MVT) we have that $g(1) - g(0) = g^{'}(c)$ for some $c$ in between 0 and 1. This implies that $g(1) = \ell(\w_1)$, $g(0) = \ell(\w_2)$ and $g^{'}(c) = \big< \nabla \ell(\w_1 t + \w_2 (1-t)), \w_1 - \w_2 \big>$. Thus from MVT we get:

\begin{align}
\nonumber |\ell(\w_1) - \ell(\w_2)| &= |\big< \nabla \ell(\w_1 t + \w_2 (1-t)), \w_1 - \w_2 \big>|\\
\nonumber&\stackrel{(a)}{\leq} \|\nabla \ell(\w_1 t + \w_2 (1-t))\| \|\w_1 - \w_2\|\\
&\stackrel{(b)}{\leq} (\text{max}_{\w \in C}\|\nabla \ell(\w)\|) \|\w_1 - \w_2\| \label{equation:lipschitz}
\end{align}
where $(a)$ follows from the Cauchy-Schwarz inequality and $(b)$ follows from the fact that $\w_1 t + \w_2 (1-t) \in C$ and $\forall \w \in C$, $ \|\nabla \ell(\w)\| \leq \text{max}_{\w \in C} \|\nabla \ell(\w)\|$.

When $\ell(\w)$ is quadratic, then we can always write $\ell(\w) = \dfrac{1}{2} (\w - \w^*)^T Q (\w - \w^*) + c_0$, where $Q = \nabla^2 \ell(\w)$ is a constant symmetric matrix and $c_0 = \ell(\w^*)$. From our definition of $\ell(\w)$ we can write:
\begin{align*}
    \text{max}_{\w \in C}\|\nabla \ell(\w)\| &= \text{max}_{\w \in C}\|Q(\w - \w^*)\|\\
    &\stackrel{(a)}{\leq} \text{max}_{\w \in C} \beta \|\w - \w^*\|\\
    &\stackrel{(b)}{\leq} \beta (\text{max}_{\w \in C} \|\w\| + \|\w^*\|)\\
    &\leq \beta(R + \|\w^*\|)
\end{align*}
where (a) follows from the definition of $\beta$ and (b) follows from the triangle inequality.
Substituting this result in \Cref{equation:lipschitz} we get:

\[|\ell(\w_1) - \ell(\w_2)| \leq \beta \big(R + \|\w^*\| \big) \|\w_1 - \w_2\|\]
\end{proof}

\begin{lemma}
Consider a function $L_{\D}(\w) = \dfrac{1}{n} \mathlarger{\sum}_{i \in \D} \ell_i(\w) + \dfrac{\mu}{2}\|\w\|^2$, where $\ell_i(\w): \R^d \rightarrow \R^+$, $\ell_i(0) \leq M$ and $\D$ is a dataset of size $n$. Let $\w^*_{\D} = \text{argmin}_{\w \in \R^{d}} L_{\D}(\w)$, then $\|\w^*_{\D}\| \leq \sqrt{\dfrac{2M}{\mu}}$.
\label{lemma-erm-norm}
\end{lemma}
\begin{proof}
\[ \dfrac{\mu}{2}\|\w^*_{\D}\|^2 \stackrel{(a)}{\leq} L_{\D}(\w^*_{\D}) \stackrel{(b)}{\leq} L_{\D}(0) \stackrel{(c)}{\leq} M \]

Thus, $\|\w^*_{\D}\| \leq \sqrt{\dfrac{2M}{\mu}}$, where $(a)$ follows from the assumption that $\ell_i(\w)$ is non-negative, $(b)$ follows from the fact that $\w_{\D}^*$ is the minimizer of $L_{\D}(\w)$ and $(c)$ follows from the assumption that $\ell_i(0) \leq M$. Note that the result is independent of $n$, thus, the empirical risk minimizers of the datasets obtained by removing a subset of samples will also lie within a $d-$dimensional sphere of radius $\sqrt{\dfrac{2M}{\mu}}$.

\end{proof}

\begin{lemma} Consider a function $L_{\D}(\w)=\dfrac{1}{n} \mathlarger{\sum}_{i \in \D} \widehat{\ell_i}(\w)$, where $\widehat{\ell_i}(\w) = \ell_i(\w) + \dfrac{\mu}{2} \|\w\|^2$, $\ell_i(\w): \R^d \rightarrow \R^+$ is strongly convex function with $\ell_i(0) \leq M < \infty$. Let $\Dtrr$ (retain set) be the dataset remaining after removing $b$ samples $\Dtrff$ (forget set) from $\D$ i.e. $\Dtrr = \D - \Dtrff$. Let  $\w^{*}_{\D} \triangleq \text{argmin}_{\w \in \R^d} L_{\D}(\w)$ and $\w^{*}_{\Dtrr} \triangleq \text{argmin}_{\w \in \R^d} L_{\Dtrr}(\w)$. Let $C = \{\w | \w \in \R^d \text{ and } \|\w\| \leq \sqrt{2M/\mu}\}$ and $G \triangleq \text{max}_{\w \in C} \|\nabla \ell(\w)\| < \infty $. Then we have that:
\[\|\w^{*}_{\D}-\w^{*}_{\Dtrr}\| \leq \dfrac{2bG}{n\mu}\]

When $\ell_i(\w)$ is also quadratic with $\beta = \text{max}_{i \in \D}\lambda_{M}(\nabla^2 \hat{\ell}_i(\w))$, the smoothness constant of $L_{\D}(\w)$, we have that $G = 2\beta\sqrt{2M/\mu}$.
\label{lemma-sensitivity}
\end{lemma}

\begin{proof}
We use the same technique as proposed in \cite{neel2020descenttodelete}.
\begin{align}
    \nonumber L_{\D}(\wremain) &= \dfrac{n-b}{n} L_{\Dtrr}(\wremain) + \dfrac{b}{n} L_{\Dtrff}(\wremain)\\
    \nonumber &\stackrel{(a)}{\leq} \dfrac{n-b}{n} L_{\Dtrr}(\wfull) + \dfrac{b}{n} L_{\Dtrff}(\wremain)\\
    \nonumber &= L_{\D}(\wfull) + \dfrac{b}{n} L_{\Dtrff}(\wremain) - \dfrac{b}{n} L_{\Dtrff}(\wfull)\\
    \label{equation:lemma4-1}
\end{align}
where, $(a)$ first inequality follows from the fact that $\wremain$ is the minimizer of $L_{\Dtrr}(\w)$. 

From \Cref{lemma-erm-norm} we know that $\|\wfull\|, \|\wremain\|,\|\w^*_{\Dtrff}\| \leq \sqrt{\dfrac{2M}{\mu}}$. Also from the definition of $\beta$ we have that $\lambda_{M}(\nabla^2 L_{\Dtrff}(\w)) \leq \beta$. Then applying \Cref{lemma:lipschitz} with $R = \sqrt{\dfrac{2M}{\mu}}$ for $L_{\Dtrff}(\w)$ we get that
\begin{equation}
    |L_{\Dtrff}(\wfull) - L_{\Dtrff}(\wremain)| \leq G\|\wfull - \wremain\|
    \label{equation:lemma4-2}
\end{equation}

From the the definition of $L_{\Dtrff}(\w)$ we know that it is a $\mu-$strongly convex function. So we have the following property:
\begin{equation}
    L_{\Dtrff}(\wremain) \geq L_{\Dtrff}(\wfull) + \dfrac{\mu}{2} \|\wfull-\wremain\|^{2} 
    \label{equation:lemma4-3}
\end{equation}

Substituting \Cref{equation:lemma4-2} and \Cref{equation:lemma4-3} in \Cref{equation:lemma4-1} we get:

\begin{align*}
  \dfrac{\mu}{2} \|\wfull-\wremain\|^{2} &\leq bG \|\wfull - \wremain\|\\
  \|\wfull-\wremain\| &\leq \dfrac{2bG}{n\mu}\\  
\end{align*}

When $\ell_i(\w)$ is also quadratic with $\beta = \text{max}_{i \in \D}\lambda_{\text{Max}}(\nabla^2 \hat{\ell}_i(\w))$, the smoothness constant, then from \Cref{lemma:lipschitz} we have $G=\beta \big(R + \|\w^*_{\D}\| \big) \leq \beta(2\sqrt{2M/\mu})$.

\begin{align*}
  \|\wfull-\wremain\| &\leq \dfrac{4b\beta\sqrt{2M}}{n\mu^{3/2}}\\  
\end{align*}

\end{proof}

\begin{lemma}
Let $\ell(\w): \R^d \rightarrow \R^+$ be a convex and $\beta-$ smooth with minimizer, $\w^* = \text{argmin}_{\w \in \R^d} \ell(\w)$. Then we have that: 
\begin{enumerate}
    \item $\ell(\w) - \ell(\w^*) \leq \dfrac{\beta}{2} \|\w - \w^*\|^2$
    \item $\|\nabla \ell(\w)\|^{2} \leq 2 \beta (\ell(\w) - \ell(\w^*))$
\end{enumerate}
\label{lemma:smoothness-general}
\end{lemma}
\begin{proof}
From the definition of $\beta-$smoothness we have that:
\begin{equation}
\ell(\w_1) \leq \ell(\w_2) + \big<\nabla \ell(\w_2), \w_1 - \w_2\big> + \dfrac{\beta}{2} \|\w_1 - \w_2\|^2
\label{equation:smoothness}
\end{equation}

Setting $\w_1 = \w$, $\w_2 = \w^*$ and using the fact that $\nabla \ell(\w^*) = 0$, we get that:

\[\ell(\w) - \ell(\w^*) \leq \dfrac{\beta}{2} \|\w - \w^*\|^2\]

For (2) we minimize \Cref{equation:smoothness} with respect to $\w_1$:

\begin{align}
    \nonumber \text{min}_{\w_1 \in \R^d} \ell(\w_1) &\leq \text{min}_{\w_1 \in \R^d} \big( \ell(\w_2) + \big<\nabla \ell(\w_2), \w_1 - \w_2\big> + \dfrac{\beta}{2} \|\w_1 - \w_2\|^2 \big)\\
    \nonumber&= \ell(\w_2) + \text{min}_{\w_1 \in \R^d} \big( \big<\nabla \ell(\w_2), \w_1 - \w_2\big> + \dfrac{\beta}{2} \|\w_1 - \w_2\|^2 \big)\\
    &\stackrel{(a)}{=} \ell(\w_2) - \dfrac{\|\nabla \ell(\w_2)\|^2}{2\beta}\\ \label{equation:smoothnes-general}
\end{align}

where $(a)$ follows from the result that $\w_1 = \w_2 - \dfrac{\nabla \ell(\w_2)}{\beta} = \text{argmin}_{\w_1 \in \R^d} \big( \big<\nabla \ell(\w_2), \w_1 - \w_2\big> + \dfrac{\beta}{2} \|\w_1 - \w_2\|^2 \big)$. 
Setting $\w_2=\w$, $\text{min}_{\w_1 \in \R^d} = \ell(\w^*)$ in \Cref{equation:smoothnes-general} and re-arranging the terms we get (2).
\end{proof}

\begin{theorem}(SGD)
Consider $L_{\D}(\w) = \dfrac{1}{n} \mathlarger{\sum}_{i \in \D} \ell_{i}(\w)$, where $\ell_i(\w): \R^d \rightarrow \R^+$ is $\beta-$smooth and $\mu-$ strongly convex. Let $\w^*=\text{argmin}_{\w \in \R^d}L_{\D}(\w)$ and $\w_i^*=\text{argmin}_{\w \in \R^d} \ell_i(\w)$. Then we have the following result after $t$ steps of SGD with batch-size $B$ and constant learning rate $\eta=\mu/{\beta}^2$:
\[\E \|\w_{t} - \w^*\|^2 \leq \Big(1 - \dfrac{\mu^2}{\beta^2} \Big)^t \|\w_{0} - \w^*\|^2  + \dfrac{2\sigma_{\ell}}{B\beta}\]
where $\sigma_{\ell}=\dfrac{1}{n} \mathlarger{\sum}_{i=1}^{n} \ell_i(\w^*) - \dfrac{1}{n} \mathlarger{\sum}_{i=1}^{n} \ell_i({\w_i^*})$
\label{theorem:sgd}
\end{theorem}
\begin{proof}
We do not use the bounded gradient assumption for the convergence of SGD and instead use the smoothness of our loss function \cite{bassily2018exponential}. This is because the training loss in our case is a quadratic function whose gradient is linear and not bounded.

Consider a mini-batch SGD update,
\[ \w_{t+1} = \w_{t} - \eta \cdot \dfrac{1}{B} \sum_{j=1}^{B}\nabla \ell_{i^{(j)}_{t+1}}(\w_t)\]
where the examples $\{i_{t+1}^{(1)}, i_{t+1}^{(2)}, \cdots i_{t+1}^{(B)}\}$ are sampled uniformly at random with replacement for all the iterations $t$.

Then by expanding $\|\w_{t+1} - \w^*\|^2$,
\begin{align*}
    \|\w_{t+1} - \w^*\|^2 &= \Big\| \w_{t} - \eta \cdot \dfrac{1}{B} \sum_{j=1}^{B}\nabla \ell_{i^{(j)}_{t+1}}(\w_t) - \w^* \Big\|^2\\
    &= \|\w_t - \w^*\|^2 - 2 \Big<\w_t - \w^*, \eta \cdot \dfrac{1}{B} \sum_{j=1}^{B}\nabla \ell_{i^{(j)}_{t+1}}(\w_t) \Big> + \Big\|\eta \cdot \dfrac{1}{B} \sum_{j=1}^{B}\nabla \ell_{i^{(j)}_{t+1}}(\w_t) \Big\|^2\\
\end{align*}
Now taking expectation over the randomness of sampling we get that, 
\begin{align}
    \nonumber
    \E \|\w_{t+1} - \w^*\|^2 &\leq 
    \E \|\w_t - \w^*\|^2 - 2 \underbrace{\E \Big<\w_t - \w^*, \eta \cdot \dfrac{1}{B} \sum_{j=1}^{B}\nabla \ell_{i^{(j)}_{t+1}}(\w_t) \Big>}_{(T_1)}\\ 
    &+ \underbrace{\E \Big\|\eta \cdot \dfrac{1}{m} \sum_{j=1}^{B}\nabla \ell_{i^{(j)}_{t+1}}(\w_t) \Big\|^2}_{(T_2)} 
\label{equation:sgd-split}
\end{align}
We will lower bound ($T_1$) and upper bound ($T_2$). Let $\xi_t = \{i_{t}^{(1)}, i_{t}^{(2)}, \cdots i_{t}^{(B)}\}$. Then for ($T_1$) we have,
\begin{align}
    \nonumber
    \E \Bigg[ \Big<\w_t - \w^*, \eta \cdot \dfrac{1}{B} \sum_{j=1}^{B}\nabla \ell_{i^{(j)}_{t+1}}(\w_t) \Big>\Bigg] &= \E_{\xi_{1} \cdots \xi_{t}} \Bigg[ \E_{\xi_{t+1}} \Big[ \Big<\w_t - \w^*, \eta \cdot \dfrac{1}{B} \sum_{j=1}^{B}\nabla \ell_{i^{(j)}_{t+1}}(\w_t) \Big> \Big| \xi_{1} \cdots \xi_{t} \Big] \Bigg]\\
    \nonumber &= \E_{\xi_{1} \cdots \xi_{t}} \Bigg[ \Big<\w_t - \w^*, \eta \cdot \dfrac{1}{B} \sum_{j=1}^{B}\E_{\xi_{t+1}} \nabla \ell_{i^{(j)}_{t+1}}(\w_t) \Big>\Big] \Bigg]\\
    \nonumber &=\E \Bigg[ \Big<\w_t - \w^*, \eta \cdot \dfrac{1}{B} \sum_{j=1}^{B} \nabla L_{\D}(\w_t) \Big> \Bigg]\\
    \nonumber &=\E \Bigg[ \Big<\w_t - \w^*, \eta \nabla L_{\D}(\w_t) \Big>\Bigg]\\
    &\stackrel{(a)}{\geq} \mu \eta \cdot \E \|\w_t - \w^*\|^2 \label{equation:sgd-crossterm}
\end{align}
where $(a)$ follows from the strong convexity of $L_{\D}$. Note that if $\ell_i(\w)$ is $\mu-$strongly convex then even $L_{\D}$ is $\mu-$strongly convex.

Using the same conditioning argument as before and the fact that each sample in a batch is sampled i.i.d. for ($T_2$) we get that:

\begin{align}
    \nonumber
    \E \Big\|\eta \cdot \dfrac{1}{B} \sum_{j=1}^{B}\nabla \ell_{i^{(j)}_{t+1}}(\w_t) \Big\|^2 &= \eta^2 \Big( \dfrac{1}{B} \E \big[ \| \nabla \ell_{i_{t+1}^{(1)}} (\w_t)\|^{2}\big] + \dfrac{B-1}{B} \E \|\nabla L_{\D}(\w_t)\|^2 \Big)\\
    \nonumber &\stackrel{(a)}{\leq} \eta^2 \Bigg( \dfrac{2\beta}{B} \Big(\E L(\w_t) - \dfrac{1}{n} \sum_{i=1}^{n} \ell_i({\w_i^*}) \Big) + \dfrac{2\beta(B-1)}{B} \Big( \E L_{\D}(\w_t) - \dfrac{1}{n} \sum_{i=1}^{n} \ell_i(\w^*) \Big)\Bigg)\\
    \nonumber &= \eta^2 \Bigg( \dfrac{2\beta}{B} \Big(\E L_{\D}(\w_t) - \dfrac{1}{n} \sum_{i=1}^{n} \ell_i{(\w^*)} + \dfrac{1}{n} \sum_{i=1}^{n} \ell_i(\w^*) - \dfrac{1}{n} \sum_{i=1}^{n} \ell_i({\w_i^*}) \Big)\\ 
    \nonumber &+ \dfrac{2\beta(B-1)}{B} \Big( \E L_{\D}(\w_t) - \dfrac{1}{n} \sum_{i=1}^{n} \ell_i(\w^*) \Big)\Bigg)\\
    \nonumber &= \eta^2 \Bigg( \dfrac{2\beta}{B} \Big(\dfrac{1}{n} \sum_{i=1}^{n} \ell_i(\w^*) - \dfrac{1}{n} \sum_{i=1}^{n} \ell_i({\w_i^*}) \Big)\\ 
    \nonumber &+ 2 \beta \Big( \E L_{\D}(\w_t) - \dfrac{1}{n} \sum_{i=1}^{n} \ell_i(\w^*) \Big)\Bigg)\\
    \nonumber &\stackrel{(b)}{\leq} \eta^2 \Bigg( \dfrac{2\beta}{B} \Big(\dfrac{1}{n} \sum_{i=1}^{n} \ell_i(\w^*) - \dfrac{1}{n} \sum_{i=1}^{n} \ell_i({\w_i^*}) \Big) + \beta^2 \E \|\w_t - \w^*\|^2 \Bigg)\\
    \nonumber &\leq \eta^2 \cdot \beta^2 \cdot \E \|\w_t - \w^*\|^2 + \dfrac{2\beta\eta^2}{B} \Big(\underbrace{\dfrac{1}{n} \sum_{i=1}^{n} \ell_i(\w^*) - \dfrac{1}{n} \sum_{i=1}^{n} \ell_i({\w_i^*})}_{\sigma_{\ell}} \Big)\\
    &= \eta^2 \cdot \beta^2 \cdot \E \|\w_t - \w^*\|^2 + \dfrac{2\beta\eta^2 \sigma_{\ell}}{B}
    \label{equation:sgd-gradients-var}
\end{align}
where $(a)$ follows from applying \Cref{lemma:smoothness-general} (1) to $\ell_{i^{(j)}_{t+1}}(\w_t)$ and $L_{\D}(\w_t)$ and $(b)$ follows from applying \Cref{lemma:smoothness-general} (2).

Now substituting \Cref{equation:sgd-crossterm} and \Cref{equation:sgd-gradients-var} in \Cref{equation:sgd-split}, we get that,

\begin{align}
    \E \|\w_{t+1} - \w^*\|^2 &\leq (1 - 2\eta\mu + \eta^2\beta^2) \E \|\w_{t} - \w^*\|^2  + \dfrac{2\beta\eta^2 \sigma_{\ell}}{B}
\label{equation:sgd-recursion-1}
\end{align}

Minimizing the coefficient with respect to $\eta$ we get $\eta^* = \dfrac{\mu}{\beta^2}$, which gives the following update equation:

\begin{align}
    \E \|\w_{t+1} - \w^*\|^2 &\leq \Big(1 - \dfrac{\mu^2}{\beta^2} \Big) \E \|\w_{t} - \w^*\|^2  + \dfrac{2\mu^2 \sigma_{\ell}}{B\beta^3}
\label{equation:sgd-recursion-2}
\end{align}

Now applying this update recursively we get:

\begin{align*}
    \E \|\w_{t} - \w^*\|^2 &\leq \Big(1 - \dfrac{\mu^2}{\beta^2} \Big)^t \E \|\w_{0} - \w^*\|^2  + \dfrac{2\sigma_{\ell}}{B\beta}
\end{align*}
\end{proof}

\begin{definition} (Forgetting: Single Request) Consider an ERM problem with $L_{\D}(\w) = \dfrac{1}{n} \mathlarger{\sum}_{i \in \D} \hat{\ell_i}(\w)$, where $\hat{\ell_i} = \ell_i(\w) + \dfrac{\mu}{2} \|\w\|^2$, $\ell_i(\w): \R^d \rightarrow \R^+$ is a quadratic convex function and $\D$ is a 
dataset of size $n$. Let $\w_{\D} = \A_{\T}(L_{\D}(\w))$ be the weights obtained after $\T$ steps of algorithm $\A$ (SGD in our case) on $L_{\D}(\w)$.  Then given a request to forget a set $\Dtrff \subset \D$ we apply the following scrubbing procedure:

\begin{equation}
    S(\w_{\D}, \D ,\Dtrff) \triangleq  \w_{\D} - \Delta \w_{\D, \Dtrff} + z
\label{equation:forgetting-single}
\end{equation}

where $\Delta \w_{\D, \Dtrff} = \A_{\tau}(\tilde{L}_{\D-\Dtrff}(\w))$, $\tau$ is the number of steps of $\A$ to minimize $\tilde{L}_{\Dtrr}(\w)$ (where $\Dtrr = \D - \Dtrff$). Here  
\begin{equation}
    \tilde{L}_{\Dtrr}(\w) = 0.5 \cdot \w^T H_{\Dtrr} \w  - \w^T g_{\Dtrr}(\w_{\D})
\label{equation:forget-problem}
\end{equation}
$z \sim \N(0, \sigma I)$ and $H_{\Dtrr}$ is the hessian on the remaining data ($\Dtrr$). We compute the \textit{residual gradient}, $g_{\Dtrr}(\w_{\D}) = \nabla L_{\Dtrr}(\w_{\D})$ once over entire $\Dtrr$, while $\w^T H_{\Dtrr}(\w_{\D})\w$ is compute stochastically in $\A$.
\label{definition:forgetting-single}
\end{definition}

\begin{definition} (Forgetting: Multiple Requests)
Consider that we are provided with a sequence of forgetting request $(\Dtrff^j)$. Let $\D_j \subset \D$ be the dataset remaining, $\w_{\D_j}$ (or simply $\w_j$) be the weights obtained after $j$ forgetting requests. Then given the ${j+1}^{\text{th}}$ request to forget $\Dtrff^{j+1} \subset \D_j$, from \Cref{equation:forgetting-single} in \Cref{definition:forgetting-single} we have that:
\begin{equation}
    \w_{j+1} \triangleq S(\w_j, \D_j, \Dtrff^{j+1}) = \w_{j} - \Delta \w_{\D_j, \Dtrff^{j+1}} + z
\end{equation}
\label{definition:forgetting-sequential}
where $z \sim \N(0, \sigma I)$ and $\w_0 = \w_\D$ are the weights obtained after training on the entire data $\D$.
\end{definition}

\begin{theorem}(Formal)
Consider an empirical risk, $L_{\D}(\w)=\dfrac{1}{n} \mathlarger{\sum}_{i \in \D} \widehat{\ell_i}(\w)$, where $\widehat{\ell_i}(\w) = \ell_i(\w) + \dfrac{\mu}{2} \|\w\|^2$, $\ell_i(\w): \R^d \rightarrow \R^+$ is quadratic convex function with symmetric $\nabla^2 \ell(\w)$, $\beta = \text{max}_{i \in \D}\lambda_{M}(\nabla^2 \hat{\ell}_i(\w))$ is the smoothness constant of $L_{\D}(\w)$, $\ell_i(0) \leq M < \infty$ and $\D$ is a dataset of size $n$. Let $\A$ be SGD with mini-batch size $B$, $\sigma_{\ell} > 0$ be some constant associated with SGD,  $\gamma = 1 - \mu^2/\beta^2$, $\tau$ be the number of steps of $\A$ performed while forgetting and $a \in (0, 1/{\gamma^{\tau}}-1)$. From \Cref{definition:forgetting-sequential}, let $\w_{j-1}, {\D}_{j-1}$ be the scrubbed weights and the dataset remaining after $j-1$ removal requests, then given a forgetting request to remove $\Dtrff^j$ $(|\Dtrff^j|=b)$ we obtain the following bound on the amount of information remaining in the weights after using the scrubbing procedure in \Cref{definition:forgetting-sequential}:
\[\boxed{I(\cup_{k=1}^{j} \Dtrff^k, S(\w_{j-1}, \D_{j-1}, \Dtrff^j)) \leq \dfrac{2\gamma^{\tau} (2 + 1/\alpha) \Bigg[\Bigg(\dfrac{8b\beta\sqrt{2M}}{n\mu^{3/2}\sqrt{\sigma}}\Bigg)^2 + d\Bigg]  +  \dfrac{8\sigma_{\ell}}{B\beta\sigma}}{1 - (1 + \alpha) \gamma^{\tau}}}\]
\label{theorem-information-formal}
\end{theorem}

\begin{proof}
We follow similar proof technique to \cite{neel2020descenttodelete}. Consider a dataset $\D$ of size $n$ and a forgetting sequence $\D_{Forget} = (\Dtrff^j)_j$ (batches of data of size $b$ that we want to forget). Let $\w_j,\w^{'}_j,\hat{\w}_j,\D_j$ be the scrubbed weights with noise, scrubbed weights without noise, weights obtained by re-training from scratch using SGD and the remaining dataset after $j$ requests of forgetting. Let $n_j$ be the size of $\D_j$ and $\w^*_{j} = \text{argmin}_{\w \in \R^d} L_{\D_j}(\w)$. Then from \Cref{definition:forgetting-sequential} we have that 
\begin{equation}
    \w_j = \w^{'}_j + z
\label{equation:forgetting-1}
\end{equation}
\begin{equation}
    \w^{'}_j = \w_{j-1} - \Delta \w_{\D_{j-1}, \Dtrff^j}
\label{equation:forgetting-2}
\end{equation}
where $z \sim \N(0, \sigma^2 I)$,  $\Delta \w_{\D_{j-1}, \Dtrff^j} = \A_{\tau}(\tilde{L}_{\D_{j-1}-\Dtrff^j}(\w))$, $\A_{\tau}$ is $\tau$ steps of SGD and $\tilde{L}_{\D_{j-1}-\Dtrff^j}(\w) = \dfrac{1}{2|\D_{j-1} - \Dtrff^j|} \sum_{i \in \D_{j-1}-\Dtrff^j} \w^T \nabla^2 \ell_{i}(\w_\D) \w - \big< \nabla L_{\D_{j-1} - \Dtrff^j}(\w_{\D}), \w \big>$. Note that $\hat{\w}_j$ are weights obtained at the end of training with SGD while $\w^*_j$ is the true empirical risk minimizer of $L_{\D_j}(\w)$.

After re-training from scratch on $\D_j$ for $\T_j$ steps using SGD with mini-batch size of $B$, we have the following relation for $j \geq 0$, using \Cref{theorem:sgd}:
\begin{align}
\E\|\hat{\w}_{j} - \w^*_{j}\|^{2} \leq \gamma^{\T_{j}} \|\hat{\w}_{\text{init}} - \w^*_{j}\|^{2} + \dfrac{2\sigma_{\ell}}{B\beta}
\label{equation:sgd-1}
\end{align}
where $\gamma = 1 - \mu^2/\beta^2$ and $\hat{\w}_{\text{init}}=0$ is the training initialization, $\hat{\w}_0$ are the weights obtained by training on $\D_0 = \D$, which is the complete dataset before receiving any forgetting request. When training the linearized model the user weights are initialized to $0$ since they correspond to the first order perturbation of the non-linear weights. 

Let us select ${\T}_{j} \geq \tau + \dfrac{2 \log{({n_{j}\mu}/{4b\beta}})}{\log{{1}/{\gamma}}}$, where $n_j \geq \dfrac{n}{2}$. Here $\tau$ is the number of steps of SGD used to remove one batch ($b$ samples) of data during forgetting. Substituting $\tau_j$ in \Cref{equation:sgd-1} we get:
\begin{equation}
    \E\|{\hat{\w}_{j}} - \w^*_{j}\|^{2} \leq
    \gamma^{\tau} \bigg(\dfrac{4b\beta}{n_j \mu}\bigg)^2 \|\hat{\w}_{\text{init}} - \w^*_{j}\|^{2} + \dfrac{2\sigma_{\ell}}{B\beta}
    \stackrel{(a)}{\leq} \gamma^{\tau} \bigg(\dfrac{8b\beta\sqrt{2M}}{n\mu^{3/2}}\bigg)^2 + \dfrac{2\sigma_{\ell}}{B\beta}
    \label{equation:retrain-bound}
\end{equation}
where (a) follows from $\|\hat{\w}_{\text{init}} - \w^*_{j}\| \leq \sqrt{\dfrac{2M}{\mu}}$ and $1/n_j \leq 2/n$.

Now we will compute a similar bound for the weights obtained by applying the forgetting procedure.
For any $j \geq 1$, using the scrubbing procedure in \Cref{definition:forgetting-sequential} we have the following relation:
\begin{equation}
    \E\|\w^{'}_{j} - \w^*_{j}\|^{2} \leq \dfrac{\gamma^{\tau} (1 + 1/\alpha)\Bigg[\bigg(\dfrac{8b\beta\sqrt{2M}}{n\mu^{3/2}}\bigg)^2 + d\sigma^2 \Bigg]  +  \dfrac{2\sigma_{\ell}}{B\beta}}{1 - (1 + \alpha) \gamma^{\tau}}
\label{equation:forget-bound}
\end{equation}
for $0 < \alpha < 1/\gamma^{\tau} - 1$. Note that $\w_j$ are the weights after applying the newton update but before adding the scrubbing noise.

From the scrubbing procedure described in \cref{equation:forgetting-1} and \cref{equation:forgetting-2} to solve for $\Delta \w_{\D_{j-1}, \Dtrff^j}$ we minimize $\hat{L}_{\D_{j-1} - \Dtrff^j}$ using SGD. While the optimal value of $\Delta \w^*_{\D_{j-1}, \Dtrff^j} = \w_{j-1} - \w^*_{j}$, thus, $\Delta \w_{\D_{j-1}, \Dtrff^j} - \Delta \w^*_{\D_{j-1}, \Dtrff^j} = \Delta \w_{\D_{j-1}, \Dtrff^j} - \w_{j-1} + \w^*_{j} = \w^*_{j} - \w^{'}_{j}$.
We can bound $\E \|\Delta \w_{\D_{j-1}, \Dtrff^j} - \Delta \w^*_{\D_{j-1}, \Dtrff^j}\|^2$ using \Cref{theorem:sgd} and thus also bound $\E \|\w^*_{j} - \w^{'}_{j}\|$.

More precisely we have:
\begin{align}
  \nonumber
  \E \|\Delta \w_{\D_{j-1}, \Dtrff^{j}} - \Delta \w^*_{\D_{j-1}, \Dtrff^j}\|^2 &\stackrel{(a)}{=} \E \|\w^{'}_j - \w^{*}_{j}\|^{2}\\ \nonumber
  \E \|\w^{'}_j - \w^{*}_{j}\|^{2} &\stackrel{(b)}{\leq} \gamma^{\tau} \E \|\Delta \w^{(0)}_{\D_{j-1}, \Dtrff^j} - \Delta \w^*_{\D_{j-1}, \Dtrff^j}\|^2 + \dfrac{2 \sigma_{\ell}}{B\beta}\\
  \E \|\w^{'}_j - \w^{*}_{j}\|^{2} &\stackrel{(c)}{=} \gamma^{\tau} \E \|\w_{j-1} - \w^*_j\| + \dfrac{2 \sigma_{\ell}}{B\beta}
\label{equation:scrubbing-recursion}
\end{align}
where the (a) follows from the definition of the scrubbing update as shown above, (b) follows from \Cref{theorem:sgd} and (c) follows from $\Delta \w^{(0)}_{\D_{j-1}, {\Dtrff}_{j}}=0$. Note that both while training (${\T}_j$ iterations) and forgetting ($\tau$ iterations) we use SGD with a constant step-size.
We will use \cref{equation:scrubbing-recursion} along with induction to prove \cref{equation:forget-bound}. For $\z \sim \mathcal{N}(0,\sigma^2 I)$, we have:
\begin{equation}
    \E \|\z\|^{2} = d\sigma^2
\label{equation:scrubbing-noise-bound}
\end{equation}
To prove \cref{equation:forget-bound} we use induction. Lets consider the base case $j=1$. 
\begin{align*}
\E\|\w^{'}_{1} - \w^*_{1}\|^{2} &\stackrel{(1)}{\leq} \gamma^{\tau} \E \|\w_{0} - \w^*_{1}\|^{2} + \dfrac{2\sigma_{\ell}^2}{B\beta}\\
&\stackrel{(2)}{=} \gamma^{\tau} \E \|\hat{\w}_{0} - \w^*_{0} + \w^*_{0} - \w^*_{1}\|^{2} + \dfrac{2\sigma_{\ell}}{B\beta}\\
&\stackrel{(3)}{\leq} \gamma^{\tau} (1 + \alpha) \E \|\hat{\w}_{0} - \w^*_{0}\|^{2} + \gamma^{\tau} (1 + 1/\alpha) \E \|\w^*_{0} - \w^*_{1}\|^{2} + \dfrac{2\sigma_{\ell}}{B\beta}\\
&\stackrel{(4)}{\leq} \gamma^{\tau} (1 + \alpha) \E \|\hat{\w}_{0} - \w^*_{0}\|^{2} + \underbrace{\gamma^{\tau} (1 + 1/\alpha) \Bigg[\Bigg(\dfrac{8b\beta\sqrt{2M}}{n\mu^{3/2}} \Bigg)^2 + d\sigma^2 \Bigg]  +  \dfrac{2\sigma_{\ell}}{B\beta}}_{(A)}\\
&\stackrel{(5)}{\leq} \gamma^{\tau} (1 + \alpha)\underbrace{\Big[\gamma^{\tau} \Bigg(\dfrac{8b\beta\sqrt{2M}}{n\mu^{3/2}}\Bigg)^2 + \dfrac{2\sigma_{\ell}}{B\beta}\Big]}_{\leq (A)} + (A)\\
&\leq {\gamma^{\tau} (1 + \alpha)}(A) + (A)\\
&\leq \dfrac{\gamma^{\tau} (1 + \alpha)}{1-(1 + \alpha)\gamma^{\tau} }(A) + (A)\\
&= \dfrac{\gamma^{\tau} (1 + 1/\alpha) \Bigg[\Bigg(\dfrac{8b\beta\sqrt{2M}}{n\mu^{3/2}} \Bigg)^2 + d\sigma^2 \Bigg]  +  \dfrac{2\sigma_{\ell}}{B\beta}}{1 - (1 + \alpha) \gamma^{\tau}}\\
\end{align*}
where (1) follows from \cref{equation:scrubbing-recursion}, (2) follows from the \Cref{definition:forgetting-sequential}, (3) follows from \Cref{lemma-young}, (4) follows from \Cref{lemma-sensitivity} and \cref{equation:scrubbing-noise-bound}, (5) follows from \cref{equation:retrain-bound}. Note that in the base case $\w_0 = \hat{\w}_0$ which are the weights obtained by training on the complete data. For \Cref{lemma-sensitivity}, $\D_j$ and $\D_{j-1}$ differ in $b$ samples and $n_j \geq n/2$. Also note that the expectation above is with respect to the randomness of SGD.

Now that we have the base case, for any general $j>1$ we get:
\begin{align*}
\E\|\w'_{j} - \w^*_{j}\|^{2} &\stackrel{(1)}{\leq} \gamma^{\tau} \E \|\w_{j-1} - \w^*_{j}\|^{2} + \dfrac{2\sigma_{\ell}}{B\beta}\\
&\stackrel{(2)}{=} \gamma^{\tau} \E \|\w^{'}_{j-1} - \w^*_{j} + \z\|^{2} + \dfrac{2\sigma_{\ell}}{B\beta}\\
&= \gamma^{\tau} \E \|\w^{'}_{j-1} - \w^*_{j-1} + \w^*_{j-1} - \w^*_{j} + \z\| + \dfrac{2\sigma_{\ell}}{B\beta}\\
&\stackrel{(3)}{\leq} \gamma^{\tau} (1 + \alpha) \E \|\w^{'}_{j-1} - \w^*_{j-1}\|^{2} + \gamma^{\tau} (1 + 1/\alpha) \E \|\w^*_{j-1} - \w^*_{j} + \z\|^{2} + \dfrac{2\sigma_{\ell}}{B\beta}\\
&\stackrel{(4)}{\leq} \gamma^{\tau} (1 + \alpha) \underbrace{\E \|\w^{'}_{j-1} - \w^*_{j-1}\|^{2}}_{\leq \dfrac{(A)}{1 - \gamma^{\tau}(1+\alpha)}} + \underbrace{\gamma^{\tau} (1 + 1/\alpha) \Bigg[\Bigg(\dfrac{8b\beta\sqrt{2M}}{n\mu^{3/2}}\Bigg)^2 + d\sigma^2 \Bigg]  +  \dfrac{2\sigma_{\ell}}{B\beta}}_{(A)}\\
&\leq \dfrac{\gamma^{\tau} (1 + \alpha)}{1-\gamma^{\tau} (1 + \alpha)}(A) + (A)\\
&= \dfrac{\gamma^{\tau} (1 + 1/\alpha) \Bigg[\Bigg(\dfrac{8b\beta\sqrt{2M}}{n\mu^{3/2}}\Bigg)^2 + d\sigma^2 \Bigg]  +  \dfrac{2\sigma_{\ell}}{B\beta}}{1 - (1 + \alpha) \gamma^{\tau}}\\
\end{align*}
where (1) follows from \cref{equation:scrubbing-recursion}, (2) follows from the \cref{equation:forgetting-1}, (3) follows from \Cref{lemma-young}, (4) follows from induction update, \Cref{lemma-sensitivity} and \cref{equation:scrubbing-noise-bound}. For \Cref{lemma-sensitivity}, $\D_i$ and $\D_{i-1}$ differ in $b$ samples. The expectation above is with respect to the randomness of SGD and the scrubbing noise $\z$.

Thus we have:
\begin{equation}
    \E\|\w'_{j} - \w^*_{j}\|^{2} \leq \dfrac{\gamma^{\tau} (1 + 1/\alpha) \Bigg[\Bigg(\dfrac{8b\beta\sqrt{2M}}{n\mu^{3/2}}\Bigg)^2 + d\sigma^2 \Bigg]  +  \dfrac{2\sigma_{\ell}}{B\beta}}{1 - (1 + \alpha) \gamma^{\tau}}
\label{equation:forget-bound}
\end{equation}

Combining \cref{equation:retrain-bound} and \cref{equation:forget-bound} using \Cref{lemma-young} we get:
\begin{align*}
\E \|\w^{'}_j - \hat{\w}_j\|^{2} &= \E \|\w^{'}_j - \w^*_{j} + \w^*_{j} - \hat{\w}_j\|_{2}^{2}\\
&\stackrel{(1)}{\leq} 2 \E \|\w^{'}_j - \w^*_{j}\|^{2} + 2 \E \|\w^*_{j} - \hat{\w}_j\|^{2}\\
&\leq \dfrac{2\gamma^{\tau} (1 + 1/\alpha) \Bigg[\Bigg(\dfrac{8b\beta\sqrt{2M}}{n\mu^{3/2}}\Bigg)^2 + d\sigma^2 \Bigg]  +  \dfrac{4\sigma_{\ell}}{B\beta}}{1 - (1 + \alpha) \gamma^{\tau}} + 2\gamma^{T} \bigg(\dfrac{8b\beta\sqrt{2M}}{n\mu^{3/2}}\bigg)^2 + \dfrac{4\sigma_{\ell}}{m\beta}\\
&\leq \dfrac{2\gamma^{\tau} (2 + 1/\alpha) \Bigg[\Bigg(\dfrac{8b\beta\sqrt{2M}}{n\mu^{3/2}}\Bigg)^2 + d\sigma^2 \Bigg]  +  \dfrac{8\sigma_{\ell}}{B\beta}}{1 - (1 + \alpha) \gamma^{\tau}} 
\end{align*}
where (1) follows from \Cref{lemma-young}.

Thus, we get:
\begin{align}
\E \|\w^{'}_j - \hat{\w}_j\|^{2} \leq \dfrac{2\gamma^{\tau} (2 + 1/\alpha) \Bigg[\Bigg(\dfrac{8b\beta\sqrt{2M}}{n\mu^{3/2}} \Bigg)^2 + d\sigma^2 \Bigg]  +  \dfrac{8\sigma_{\ell}}{B\beta}}{1 - (1 + \alpha) \gamma^{\tau}}
\label{equation:forget-retrain-bound}
\end{align}

Since the problem is quadratic, the hessian will be same at all points. From Proposition 1 in \cite{golatkar2020forgetting}, we have:
\begin{align*}
    I(\cup_{k=1}^{j} \Dtrff^k, S(\w_{j-1}, \D_{j-1}, \Dtrff^j)) &\leq \E \big[(\w'_j - \hat{\w}_j)^T (\sigma^2 I)^{-1} (\w'_j - \hat{\w}_j) \big]\\
    &\leq \dfrac{\E \|\w'_i - \hat{\w}_i\|^{2}_{2}}{\sigma^2}
\end{align*}

Note that the expectation in the previous expression is with respect to the randomness in the training algorithm and the forgetting algorithm, plus the set $\Dtrff^k$. Then using \cref{equation:forget-retrain-bound} with the previous equation we obtain that:

\begin{equation}
    I(\cup_{k=1}^{j} \Dtrff^k, S(\w_{j-1}, \D_{j-1}, \Dtrff^j)) \leq \dfrac{2\gamma^{\tau} (2 + 1/\alpha) \Bigg[\Bigg(\dfrac{8b\beta\sqrt{2M}}{n\mu^{3/2}\sigma}\Bigg)^2 + d\Bigg]  +  \dfrac{8\sigma_{\ell}}{B\beta\sigma^2}}{1 - (1 + \alpha) \gamma^{\tau}}
    \label{equation:remaining-information}
\end{equation}

\end{proof}

\end{document}